\documentclass[pdflatex,sn-mathphys-num]{sn-jnl}

\usepackage{graphicx}%
\usepackage{multirow}%
\usepackage{amsmath,amssymb,amsfonts}%
\usepackage{amsthm}%
\usepackage{mathrsfs}%
\usepackage[title]{appendix}%
\usepackage{xcolor}%
\usepackage{textcomp}%
\usepackage{manyfoot}%
\usepackage{booktabs}%
\usepackage{makecell}%
\usepackage{algorithm}%
\usepackage{algorithmicx}%
\usepackage{algpseudocode}%
\usepackage{listings}%
\usepackage{hyperref}

\theoremstyle{thmstyleone}%
\theoremstyle{thmstyletwo}%

\theoremstyle{thmstylethree}%

\DeclareMathOperator*{\argmax}{arg\,max}
\definecolor{darkgreen}{rgb}{0.0, 0.5, 0.0}
\newcommand{\green}[1]{\textcolor{darkgreen}{#1}}
\newcommand{\red}[1]{\textcolor{red}{#1}}

\begin{document}

\title[Self-Supervision Enhances Instance-based Multiple Instance Learning Methods in Digital Pathology: A Benchmark Study]{\textbf{Self-Supervision Enhances Instance-based Multiple Instance Learning Methods in Digital Pathology: A Benchmark Study}}

\author*[1,2]{\fnm{Ali} \sur{Mammadov}}\email{ali.mammadov@ip-paris.fr}

\author[1]{\fnm{Loïc} \sur{Le Folgoc}}

\author[2]{\fnm{Julien} \sur{Adam}}

\author[2]{\fnm{Anne} \sur{Buronfosse}}

\author[2]{\fnm{Gilles} \sur{Hayem}}

\author[2]{\fnm{Guillaume} \sur{Hocquet}}

\author[1]{\fnm{Pietro} \sur{Gori}}

\affil[1]{\orgdiv{Telecom Paris}, \orgname{Institut Polytechnique de Paris}, \orgaddress{ \country{France}}}

\affil[2]{\orgname{Groupe Hospitalier Paris Saint-Joseph}, \orgaddress{\country{France}}}


\abstract{Multiple Instance Learning (MIL) has emerged as the best solution for Whole Slide Image (WSI) classification. It consists of dividing each slide into patches, which are treated as a bag of instances labeled with a global label. MIL includes two main approaches: instance-based and embedding-based. In the former, each patch is classified independently, and then the patch scores are aggregated to predict the bag label. In the latter, bag classification is performed after aggregating patch embeddings. Even if instance-based methods are naturally more interpretable, embedding-based MILs have usually been preferred in the past due to their robustness to poor feature extractors. However, recently, the quality of feature embeddings has drastically increased using self-supervised learning (SSL). Nevertheless, many authors continue to endorse the superiority of embedding-based MIL. To investigate this further, we conduct 710 experiments across 4 datasets, comparing 10 MIL strategies, 6 self-supervised methods with 4 backbones, 4 foundation models, and various pathology-adapted techniques. Furthermore, we introduce 4 instance-based MIL methods never used before in the pathology domain. Through these extensive experiments, we show that with a good SSL feature extractor, simple instance-based MILs, with very few parameters, obtain similar or better performance than complex, state-of-the-art (SOTA) embedding-based MIL methods, setting new SOTA results on the BRACS and Camelyon16 datasets. Since simple instance-based MIL methods are naturally more interpretable and explainable to clinicians, our results suggest that more effort should be put into well-adapted SSL methods for WSI rather than into complex embedding-based MIL methods.}


\maketitle
\section{Introduction}
\label{sect:intro}  
Whole Slide histopathology Image (WSI)  {analysis has become an increasingly common tool} for disease diagnosis in digital pathology \cite{gurcan_histopathological_2009} . However, the gigapixel size of WSIs, makes the manual analysis very time-consuming and presents significant challenges for conventional Deep Learning (DL) methods~\cite{cheplygina_not-so-supervised_2019, srinidhi_deep_2021}, as they are not designed to support such large images. To address that, a simple approach involves dividing the WSI into smaller patches that DL methods can easily handle. Then, features or predictions from a patch-level encoder/classifier are aggregated to get the slide-level prediction \cite{hou_patch-based_2016, wei_pathologist-level_2019}. Nonetheless, this method requires very expensive patch-level annotations, which are not always available. Please note that a naive assignment of the slide label to all patches might be clinically incorrect, since the tissue section characterizing a disease might only occupy a small fraction of the slide, while all other patches should be considered as healthy.
\begin{figure*}[t!]
\centering
\includegraphics[width=\textwidth]{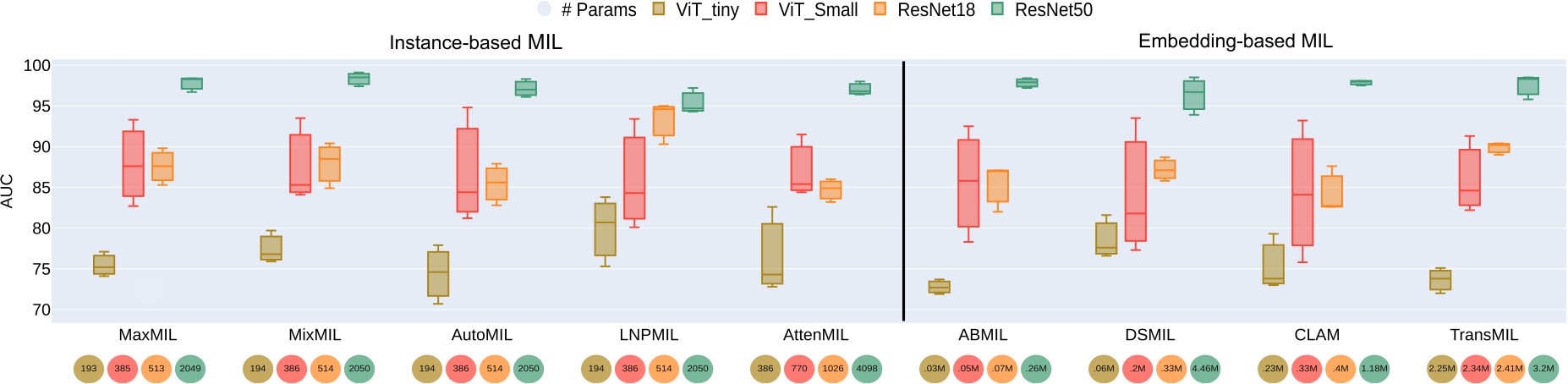}
  \caption{\textit{Self-supervision (SSL) closes the gap between instance- and embedding-based multiple-instance learning (MIL) methods in whole slide image (WSI) classification.} Instance-based MIL methods, when combined with robust SSL feature extractors, are on par or even outperform complex, state-of-the-art embedding-based methods in WSI classification. The Y-axis represents the AUC scores of different MIL methods for four backbones (Vit\_tiny, ViT\_Small, ResNet18 and ResNet50) on the Camelyon16 dataset (at a resolution 20x for ResNet50 and 10x for the other backbones). For each backbone and MIL method, we show the box-plot of 3 self-supervised pre-trainings: DINO \cite{caron_emerging_2021}, MOCO-V3 \cite{he_momentum_2020} and Barlow Twins \cite{zbontar_barlow_2021}.
}
  \label{fig:intro}
\end{figure*}
\subsection{Multiple-Instance Learning (MIL)} To tackle this challenge, Multiple-Instance Learning (MIL) methods, coupled with Deep Learning Feature Extraction (FE), have emerged for WSI classification using only slide (i.e., weak) labels. This approach is considered the most prominent solution in the field of histopathology, where FE avoids the costly and experience-based feature engineering part, while MIL eliminates the need for patch-level (or pixel-level) annotations \cite{tschuchnig_evaluation_2022, qu_towards_2022}. Under MIL formulation, each WSI is treated as a "bag" containing multiple instances in the form of patches, which are embedded with Convolutional Neural Network (CNN) or Vision Transformer (ViT) \cite{dosovitskiy_image_2021} backbones. The bag is labeled positive (i.e., diseased) if at least one of its patches is positive, or negative if all patches are negatives \cite{campanella_clinical-grade_2019, ilse_attention-based_2018}. In general, the existing methodologies follow a two-step pipeline: 1) feature extraction from individual patches, and 2) MIL aggregation through a pooling operation to predict the slide label \cite{li_graph_2018, lu_data-efficient_2021, li_dual-stream_2021, shao_transmil_2021, zhang_dtfd-mil_2022, guan_node-aligned_2022}. 
\noindent Many works in the literature focus on the second part of the pipeline,  developing various aggregation methods that can be categorized into two groups: \textit{instance-based} and \textit{embedding-based} methods. Instance-based methods use an instance-level classifier, which predicts a score for each patch. Then, these scores are aggregated via a MIL-based pooling operator to make the final prediction for the entire slide. 
Common pooling operators include average-pooling (\textit{MeanMIL}) and max-pooling (\textit{MaxMIL}) \cite{wang_comparison_2019, bieder_comparison_2021}.
Instance-based MIL methods are actively used in several domains \cite{carbonneau_multiple_2018, gonthier_multiple_2022}, and in particular in sound event detection \cite{wang_comparison_2019, mcfee_adaptive_2018}. However, these methods were only used in few and early works in the field of Digital Pathology, such as in \cite{campanella_clinical-grade_2019}. Even if these methods are naturally interpretable and easily explainable, they highly depend on the quality of the embedding. To increase reliability, researchers proposed to aggregate features instead of scores, moving the classification head after the pooling. These are called embedding-based methods. 

The existing literature has proposed several approaches for feature aggregation based on: deep self-attention mechanism \cite{lu_data-efficient_2021, ilse_attention-based_2018}, graph convolutional networks  \cite{li_graph_2018, chen_whole_2021, zhao_predicting_2020}, clustering \cite{sharma_cluster--conquer_2021, guan_node-aligned_2022}, transformer \cite{shao_transmil_2021} and additional training \cite{zhang_dtfd-mil_2022, lin_interventional_2023}. These pooling mechanisms are usually more complex (more parameters) than instance-based ones. On the one hand, this means that the model has a greater capability of learning how to correctly aggregate the features and thus might have a greater prediction power. On the other hand, interpretability and explainability can decrease\footnote{even though attention and self-attention mechanisms can be exploited in that regard} and at the same time computational complexity and overfitting can increase (i.e., the number of needed training samples increases).\\
\noindent Until recently, most of the researchers used ImageNet pre-trained models to extract features. However, as shown in \cite{matsoukas_what_2022,raghu_transfusion_2019}, these models might not be optimal for histopathology images and may produce poor representations (i.e., features) due to the domain gap between natural and medical images.  
This might explain why most of the early, ImageNet-based works reported that embedding-based MIL models outperformed instance-based ones  \cite{hashimoto_multi-scale_2020, kanavati_weakly-supervised_2020, wang_revisiting_2018,shao_transmil_2021}. In recent years, to enhance feature quality, the advent of large, weakly annotated WSI datasets has led many researchers to investigate self-supervised learning methods.

\subsection{Self Supervised Learning (SSL)} SSL is a paradigm whose goal is to learn useful and relevant representations for downstream tasks by leveraging pretext tasks from unlabeled data. It has shown promise in improving the performance of image classification \cite{grill_bootstrap_2020, he_momentum_2020, he_masked_2022, chen_simple_2020, zbontar_barlow_2021, caron_emerging_2021}. In digital pathology, SSL has been actively used in recent works. Notably, in  \cite{li_dual-stream_2021} authors use SimCLR \cite{chen_simple_2020} to train a feature extractor. Other authors use MoCo \cite{he_momentum_2020} in \cite{dehaene_self-supervision_2020, saillard_self-supervised_2021}, DINO \cite{caron_emerging_2021} in \cite{chen_self-supervised_2021}, BYOL \cite{grill_bootstrap_2020} in \cite{wang_transpath_2021} and MoCo V3 \cite{chen_empirical_2021} in \cite{wang_scl-wc_2022}. Some recent papers have even proposed novel self-supervised learning methods specifically tailored for digital pathology and achieved new state-of-the-art results. In \cite{chen_scaling_2022}, authors apply the DINO method at different scales for extracting hierarchical features, providing a more comprehensive representation of the histopathology images. Another significant contribution is the Semantically Relevant Contrastive Learning (SRCL) method introduced by \cite{wang_transformer-based_2022}. Based on MoCoV3, SRCL additionally compares relevance between instances to mine more positive pairs. An original self-supervised method based on an auto-encoder was explored by \cite{boyd_self-supervised_2021}, which learns to reconstruct the masked borders of an image from a given center.  In \cite{wu_improving_2023}, authors enhance the BYOL method by incorporating a clustering constraint. Additionally, \cite{lu_semi-supervised_2019} and \cite{lu_smile_2021} propose different and innovative contrastive learning-based approaches. Another line of research focuses on the use of SSL at the slide level, as proposed by \cite{lazard_giga-ssl_2023, tang_multiple_2023}. All of these approaches have shown that self-supervision outperforms ImageNet pre-training on both patch-level and slide-level downstream tasks.

\subsection{Existing Limitations} Even if recent works provided very insightful results for WSI classification using SSL methods, their experiments and comparison present some limitations. They either use a single SSL method~\cite{wang_transpath_2021,liu_multiple_2023}, or compare few SSL methods \cite{chen_self-supervised_2021}, or multiple SSL methods but with \textit{different} backbones \cite{kang_benchmarking_2023}, or they use a \textit{single} backbone for all methods \cite{shao_transmil_2021}. Furthermore, most of these works focus on a few embedding-based MIL methods, using at most one/two instance-based methods \cite{wu_improving_2023,dehaene_self-supervision_2020,shao_transmil_2021,liu_multiple_2023}. Another limitation is that some works use few datasets (and usually with a low clinical complexity\footnote{clinical complexity is measured in terms of average AUC score.}, like Camelyon16) for binary classification only \cite{wu_improving_2023}.\\
\noindent To address these limitations, we present here a fair, complete and thorough comparison on \textit{four} different datasets, for both binary and multi-class classification, that present an increasing level of clinical complexity. We investigate the importance of all the key elements of the WSI classification pipeline, testing: 4 different \textit{feature extractors (backbones)}, 6 \textit{SSL} strategies (plus a pre-training of Image-Net), 4 \textit{foundation models} and 10 \textit{MIL} methods. This research fills a significant gap in the literature of WSI classification by providing a comprehensive study and showcasing the potential of simple and interpretable instance-based MIL methods, when combined with robust SSL methods.

\subsection{Our Contributions and Main Results}
\noindent - We conduct a large-scale study on 4 datasets, with 7 different pre-training methods, 4 \textit{foundation models}, 4 different backbone models, and 10 multiple-instance learning methods for a total of 710 different configurations.\\
-  We propose simple instance-based MIL approaches based on pooling mechanisms borrowed from the sound event detection field which, to the best of our knowledge, have never been used in pathology. \\
- We demonstrate that simple instance-based MIL methods, when combined with robust SSL feature extractors, are on par or even outperform complex embedding-based methods in WSI classification, \textit{whatever} the employed backbone model (CNN or ViT), see Fig.\ref{fig:intro}.\\
- Using our newly proposed instance-based MIL methods, we achieve new  {state-of-the-art} (SOTA) results in BRACS \cite{wang_scl-wc_2022} and Camelyon16 \cite{lin_interventional_2023}, and we are on par with current SOTA methods in TCGA-NSCLC \cite{qu_boosting_2023}.\\
We also share our code, pre-trained models, extracted features, hyperparameters and valuable insights for WSI classification in \url{https://github.com/mammadov7/wsi_classification_pipeline}.

\section{Methods}
\begin{figure}[ht!]
    \centering
\includegraphics[width=0.9\textwidth]{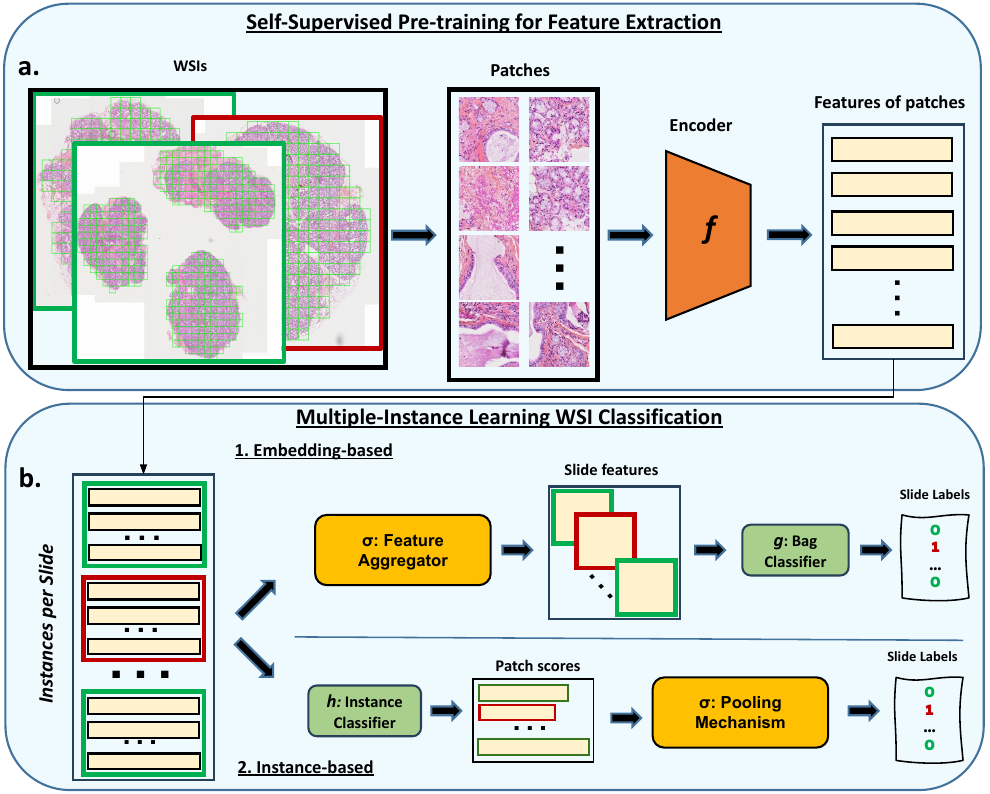}
  \caption{Pipeline of the WSI classification. At the top, a) represents a self-supervised pre-training of an encoder \textit{\textbf{f}} on patches from all slides. Once pre-trained, the encoder \textit{\textbf{f}} extracts a vector of features (i.e., representation) per patch. Representations of patches of the same slide are then stacked together. At the bottom, b.1) Embedding-based MIL: features from the same slide are aggregated together using a feature aggregator \textit{\textbf{$\sigma$}} into a bag-level representation. A bag/slide  classifier \textit{\textbf{g}} is then trained on bag/slide representations to predict the slide label. b.2) Instance-based MIL:  an instance classifier \textit{\textbf{h}} assigns scores (\textit{i.e.,} class probability) to each patch, which are then pooled using a pooling mechanisms \textit{\textbf{$\sigma$}} to predict the slide label.}  
  \label{pipeline}
\end{figure}

\subsection{MIL problem formulation} Let $X_i=\{x_{ij}\}_{j=1\dots K_i}$ be the whole slide image of subject $i$ (i.e., bag) and $x_{ij} \in \mathbb{R}^d $ its $j$-th patch (i.e. instance). We have $N \in \mathbb{Z}^+$ bags in our training set and $K_i \in \mathbb{Z}^+$ instances for each bag $i$ (here we assume that $K_i$ can vary across bags). There is a single classification label $y_i$ for each bag (binary or multi-class) and we assume that each instance has also its own label $y_{ij}$ (same classes as for the bag), but it is unknown during training.\\ 
Following \cite{ilse_attention-based_2018,javed_additive_2022}, there are two main approaches: instance-level and embedding-level. We define here a generic framework (see Fig.\ref{pipeline}) well adapted to both. Let $f(\cdot): \mathbb{R}^d \rightarrow \mathbb{R}^p$ be a parametric function, here a deep learning encoder, that maps the instances to a feature space and $\mathop{\sigma}$ a MIL pooling operator (i.e., a symmetric, permutation-invariant function). The final bag-score $S(X)$ is:

\begin{equation}
\quad \quad \quad S(X_i)=g(\mathop{\sigma}(h(f(x))))
\end{equation}

\noindent where $g(\cdot)$ and $h(\cdot)$ are bag-level and instance-level classifiers respectively. In the instance-level approach, we only use $h(\cdot)$, which computes a score (class probability) for each instance, fixing $g(\cdot)$ to the identity function. Instead, in the embedding-level approach, we use $g(\cdot)$, which computes a score at the bag level, setting $h(\cdot)$ to the identity mapping.

\subsection{ Instance-based MIL pooling operators} 
In this class of methods, the pooling mechanism is used to aggregate patch scores. The two most used MIL pooling operators $\sigma$ are the $\max(\cdot)$, \textit{MaxMIL}, and the $\text{mean}(\cdot)$, \textit{MeanMIL}, which have served as baseline in many WSI classification works, such as \cite{ilse_attention-based_2018,li_dual-stream_2021,zhang_dtfd-mil_2022,shao_transmil_2021,lerousseau_sparseconvmil_2021,javed_additive_2022,saillard_self-supervised_2021}.
However, Max-pooling may not capture sufficient information from a single patch and Average-pooling may gather excessive information from all patches, leading to suboptimal results. Unfortunately, one does not necessarily know \textit{a priori} which pooling mechanism would be the most suitable for a given dataset. In the sound event detection domain, authors have proposed simple generalizations~\cite{bieder_comparison_2021}, notably: Learned-Norm Pooling (LNP) \cite{calders_learned-norm_2014} , Mix-pooling \cite{lee_generalizing_2016}, Auto-pooling \cite{mcfee_adaptive_2018} and Attention-pooling \cite{wang_comparison_2019}. These approaches involve learning a set of weights or norms associated with each patch's contribution. The pooling mechanism is not fixed but is adaptively adjusted during the training process based on the significance of each patch for the classification task. This adaptability allows the model to assign more weight to informative patches and less to less relevant ones.
 {\subsection{Proposed Instance-based MIL pooling operators}} 
Let $y_k=h(f(x_k)) \in [0,1]$ the predicted probability (score) of patch $k$, we define:\\
\textbf{MixMIL} (Mixed-Pooling) \cite{lee_generalizing_2016}:   $\sigma = \alpha\max_{ k=1 \dots K} y_k + (1-\alpha)\sum_{k=1}^{K}y_k$ where $\alpha \in [0,1]$ is a trainable parameter.\\
\textbf{AutoMIL} (Auto-pooling) \cite{mcfee_adaptive_2018}:   $\sigma = \sum_{k=1}^{K}y_k(\frac{exp(\alpha \cdot y_k)}{ \sum_{j=1}^{K} exp(\alpha \cdot y_j)})$ where $\alpha \in [0,\infty)$ is a trainable parameter.\\
\textbf{LNPMIL} (Softmax-pooling) \cite{calders_learned-norm_2014}: $\sigma = \left(\frac{1}{K}\sum_{k=1}^{K}|y_k|^p \right)^{\frac{1}{p}}; \quad  p = 1 + \log(1+e^{\Tilde{p}})$ where $\Tilde{p} \in \mathbb{R}$ is a trainable parameter.\\ 
\textbf{AttenMIL} (Attention-pooling) \cite{wang_comparison_2019}: $\sigma = \sum_{k=1}^{K} y_k \hat{w}_k$ with $\hat{w}=$softmax($w_k$) where the weights $w_k$, one for each patch, are learned with a dedicated layer in the network. 
\subsubsection{Multi-class Instance-based MIL}
Here, we propose a simple extension to multiple classes of the MaxMIL pooling operator.\\
\textbf{Multi-class MaxMIL}: for each instance $x_{ik}$, the instance-level classifier outputs a score vector $h(x_{ik})$ whose components are the (scalar) scores $s_{ik}^{(c)} \in [0,1]$ for each class $c$. Assuming 3 classes, we have: \[ h(x_{ik}) = \left( s_{ik}^{(1)}, s_{ik}^{(2)}, s_{ik}^{(3)} \right) \] 
where we make the hypothesis that $c=1$ refers to the normal/benign class while $c=2,3$ refer to two different tumoral/pathological classes (as in BRACS \cite{brancati_bracs_2022}). Cancerous slides may exhibit large portions of normal/benign tissues, which is usually easier to classify. As a result, they will be probably assigned a high score $s_{ik}^{(1)}$ and thus, simply applying max pooling over the patch scores of all classes, it would probably select a normal tissue patch even in a cancerous slide. To avoid that, we propose to perform the aggregation by first selecting the instance with the highest score among the second and third classes: 
\[ \text{m} = \arg\max_{k}(s_{ik}^{(2)}, s_{ik}^{(3)}) \] 
\noindent Then, the bag-level prediction is made by taking the maximum score among all the scores of the instance $m$:
    \[ y_i = \arg\max  \left( s_{im}^{(1)}, s_{im}^{(2)}, s_{im}^{(3)} \right) \] 
\noindent In BRACS, the three classes are: Benign Tumor, Atypical Tumor,
and Malignant Tumor.\\ 
\textbf{Multi-class MixMIL} (Mixed-Pooling), we use the aforementioned strategy only for the MaxMIL part of MixMIL.
\subsection{ Embedding-based MIL pooling operators} 
Differently from instance-based MIL methods, here the pooling mechanisms aggregate features, and not scores (i.e., scalars). Embedding-based MIL methods are usually more complex (i.e., more parameters) since they need to effectively capture and aggregate information from high-dimensional vectors~\cite{lu_data-efficient_2021, li_dual-stream_2021, ilse_attention-based_2018, li_graph_2018, guan_node-aligned_2022, shao_transmil_2021, zhang_dtfd-mil_2022, lin_interventional_2023}. 
 In our study, we employ the following four and frequently used state-of-the-art methods:: \\
\textbf{ABMIL}~\cite{ilse_attention-based_2018}: is an attention-based MIL. It employs trainable attention layers to attribute an attention score to each patch representation, which are then aggregated via a weighted sum for the final slide classification.
\begin{equation}
\sigma = \sum_{k=1}^{K} a_k f_k  \text{ where } f_k=f(x_k) \text{ and } a_k= \frac{\exp(w^T \tanh(V f_k^T))}{\sum_{j=1}^{K} \exp(w^T \tanh(V f_j^T))}    
\end{equation}
 The parameters $w$ and $V$ are learned with trainable attention layers. \\
\textbf{CLAM}~\cite{lu_data-efficient_2021}:  also uses an attention-based aggregation mechanism, and proposes instance-level clustering over the patches to improve the feature space.\\
\textbf{DSMIL}~\cite{li_dual-stream_2021}: is an hybrid, dual-stream method that jointly learns an instance and a bag classifier. First, it detects the critical patch $m=\underset{x}{\argmax} (h(f(x_k)))$ as in MaxMIL. Then, it transforms each instance $f_k=f(x_k)$ into two vectors: query $q_k=W_q f_k$ and information $v_k=W_vf_k$, where $W_q$ and $W_v$ are two weight matrices. Query vectors are used to compute distances between every patch and the critical one: $U(f_k,f_m)=$softmax($q_k \cdot q_m$), where $\cdot$ denotes the inner product. Similarly to ABMIL, the bag aggregator $\sigma$ is a weighted sum. However, now the weights are based on the similarity of the features  {with respect to} the critical sample: $\sigma=\sum_k U(f_k,f_m)v_k$. It is used to compute the final score $S(X)=\frac{1}{2}(h(f(x_m)) + g(\sigma))$, which is an average between the instance and bag classifiers and where $h$ and $g$ are linear classifiers.\\
\textbf{TransMIL}~\cite{shao_transmil_2021}: is a transformer-based MIL that considers the correlation between different patches  (tokens) using Transformers to investigate morphological and spatial details and it uses self-attention mechanisms for aggregation. In particular, $\sigma$ is a module, called TPT module, with two Transformer layers and a position encoding layer, where Transformer layers are designed for aggregating morphological information and a Pyramid Position Encoding Generator (PPEG) is designed for encoding spatial information. \\
 {\textbf{DAMIL:} \cite{yao2020whole} is Deep Attention Multiple Instance Survival Learning. It leverages a siamese Multiple Instance Fully Convolutional Network (MI-FCN) to learn features from phenotype clusters derived from WSIs. The framework employs attention-based MIL pooling to aggregate these features into patient-level representations, enhancing interpretability and performance. \\
\textbf{DTFDMIL:} \cite{zhang_dtfd-mil_2022} is Double-Tier Feature Distillation Multiple Instance Learning. This method introduces the concept of "pseudo-bags" which are smaller subsets of patches created from the original slides. The framework employs a two-tier MIL model: the first tier processes these pseudo-bags to distill key features, while the second tier aggregates these features to make predictions at the slide level.}

\subsection{Self Supervised Pre-training of $f$} The  encoder $f(\cdot)$ is usually pre-trained using self-supervision (SSL). Different strategies exist, including contrastive learning, clustering, knowledge distillation, and information maximization methods. Among the existing SSL methods, we opt for the following six, highly-used and state-of-the-art approaches: \\
\textbf{SimCLR} \cite{chen_simple_2020} learns representations by maximizing agreement between differently augmented views of the same data sample and by minimizing agreement between different samples, via a contrastive loss in the latent space. It comprises four key components: a data augmentation module, a base encoder (\textit{e.g.,} ResNet), a projection head (\textit{e.g.,} multi-layer perceptron (MLP)), and a contrastive loss function (InfoNCE loss). Within this framework, every image in the batch undergoes a sequential process. First, it is transformed twice (i.e., augmented) to create two distinct augmented views. These augmented views are then encoded using the base encoder and projected into a latent space. Subsequently, the contrastive InfoNCE loss is applied to encourage the attraction of positive pairs (representations originating from the same image) and the repulsion of negative pairs (representations of different images) within the latent space. Upon the completion of training, we keep the encoder for downstream tasks. \\
\textbf{BYOL} \cite{grill_bootstrap_2020} uses both online and target neural networks. Its architecture is asymmetrical, with the online network consisting of three stages: encoder, projector, and predictor. The target network comprises just two stages, the encoder, and the projector, but with different weights. In this approach, only the online network is trained and the target network parameters are updated by an exponential moving average of the online parameters. During training, based on an augmented view of an image, the online network is trained to predict the target network representation of the same image under a different augmented view. For a given image, BYOL will produce two different augmented views, v1 and v2. First, v1 will pass through the online network and v2 through the target network creating r1, r2. Then, vice-versa, to symmetrize the pipeline, v2 passes through the online network and v1 through the target, resulting in r3, r4. Finally, it minimizes a loss $L=sim(r1, r2) + sim(r3, r4)$, where $sim$ is a $l_2$-normalized mean squared error function. At test time, only the encoder of the online network is used for feature extraction. \\
\textbf{Barlow Twins} \cite{zbontar_barlow_2021} share similarities with \textit{SimCLR} in that it learns representations of images under different augmented views passing through the same network. The originality of this approach is the non-contrastive loss, which requires no negative pairs or large batches. It measures the cross-correlation matrix between the embeddings obtained from a network fed with distorted versions of the same input image, and tries to make this matrix close to the identity. It makes the embedding invariant to the distortions applied, so that the outputs of the network are as similar as possible for the distorted views of the same input while ensuring that the individual components of these embeddings are not redundant. \\ 
\textbf{DINO} \cite{caron_emerging_2021}, similarly to \textit{BYOL}, employs the teacher-student distillation paradigm with two parallel networks and an update mechanism with exponential moving averages. The main differences are that it uses a ViT backbone and Cross Entropy metric as a training loss. Moreover, DINO generates a set of \textit{local} views, comprising several augmented small crops (each less than 50\% of the image area), and \textit{global} views (two large crops, each covering more than 50\% of the image). All crops pass through the student network, while only the global views pass through the teacher network. Lastly, to avoid collapse in the presence of a momentum teacher, it applies the centering and sharpening to the momentum teacher outputs. Centering prevents one dimension to dominate but encourages collapse to the uniform distribution, while sharpening has the opposite effect. \\
\textbf{MoCo v3} \cite{chen_empirical_2021} combines \textit{BYOL}'s teacher-student distillation architecture with \textit{SimCLR}'s contrastive learning and \textit{DINO}'s Vision Transformer backbone, enhancing representation learning by leveraging the best of all three approaches. \\
\textbf{MAE} \cite{he_masked_2022} is a masked autoencoder that randomly masks 75\% of the input and then learns to reconstruct the original image from the unmasked parts of the image. It uses an architecture with an encoder operating on unmasked patches and a lightweight decoder reconstructing the full image using latent representations and mask tokens. This efficient design allows for training large encoders with reduced computing and memory, where only a lightweight decoder handles the complete set of patches.  MAE uses as optimization loss the mean squared error (MSE) between the reconstructed and original images.
\subsection{Pathology Adapted SSL Methods}
In the literature, there are very few SSL methods conceived and developed specifically for histopathology images. Furthermore, they either propose well-adapted augmentations or slight modifications of existing methods (which have been mainly developed for natural images). In this section, we describe three recent works that characterize well this class of methods.\\
\textbf{PathAug} \cite{kang_benchmarking_2023}, in this work authors proposed three novel image augmentation techniques, specifically developed for histopathological images. The first technique involves vertical flipping of the input images. The second technique uses different fields-of-view by taking patches from various magnifications of the WSIs. The third technique is based on stain augmentation and normalization (RandStainNA), as proposed by \cite{shen2022randstainna}. \\
\textbf{SRCL} \cite{wang_transpath_2021} ( semantically-relevant contrastive learning ) draws inspiration from two well-known contrastive learning methods: SimCLR \cite{chen_simple_2020} and MoCo v3 \cite{chen_empirical_2021}. Authors propose to  select additional semantically relevant positive pairs using a cosine similarity metric, thereby increasing the diversity of positive samples. This approach uses three parallel paths: online, target, and shared target branches, all employing the CTransPath architecture. A memory bank, updated iteratively, stores features from the shared target branch. The online branch is learned via gradient-based optimization, while the target branch is updated through an exponential moving average of the online network parameters. Each input sample generates three views (two augmented, one original), which are passed through the three branches, with the online branch features serving as anchor. By calculating cosine similarities between the anchor and the features in the memory bank, they identify top matches as new positive samples, enhancing the contrastive loss function by maximizing the similarity between these samples and the anchor. \\
\textbf{CluBYOL} \cite{wu_improving_2023} Cluster Bootstrap Your Own Latent (CluBYOL) is a self-supervised learning approach. It extends the original BYOL \cite{grill_bootstrap_2020} framework to better capture the underlying structure of clustered data. The main idea is to integrate clustering into the self-supervised learning process, as in SwAV \cite{caron_unsupervised_2020}, allowing the model to learn more meaningful and discriminative features by leveraging the natural grouping within the data, which is adapted for different tissue types in pathology.

\subsection{Foundation Models}
Foundation models are large models that are pre-trained on large-scale datasets and can be adapted for a wide range of downstream tasks. For our study, we included one general vision model and three pathological foundation models, as zero-shot feature encoders $f(\cdot)$.\\
\textbf{DINOv2} proposed by \cite{oquab_dinov2_2023} has emerged as an improved version of the original DINO model. DINOv2 combines the loss functions of DINO \cite{caron_emerging_2021} and iBOT \cite{zhou2021ibot} with the centering of SwAV \cite{caron_unsupervised_2020}. We use the ViT\_Small model, which is pre-trained on 142 million images. \\
\textbf{PathAugFM}, proposed by \cite{kang_benchmarking_2023}, involves pre-training the ViT\_Small model over 200 epochs on a dataset containing 19 million pathological patches. The authors utilize the DINO framework coupled with pathology-adapted augmentations, which are detailed in the previous section. \\
\textbf{CTransPath} from \cite{wang_transpath_2021}, uses a hybrid backbone.  This architecture is obtained by replacing the patch embedding component of Swin Transformers with a CNN layer. \textit{CTransPath} is trained on a dataset consisting of 15 million pathological patches with the SRCL framework. \\
\textbf{UNI}, as introduced by \cite{chen2024uni}, represents a general-purpose self-supervised model tailored for pathology applications. UNI's training involves pre-training the ViT\_Large architecture with the DINOv2 framework, using a vast dataset comprising more than 100 million tissue patches. These patches are extracted from a collection of over 100,000 diagnostic hematoxylin and eosin-stained WSIs, covering a diverse array of 20 major tissue types.

\section{Experiments}
\subsection{Datasets and Experimental Setup} We evaluate the proposed MIL and SSL methods on 4 different histopathological datasets -- Camelyon16 \cite{ehteshami_bejnordi_diagnostic_2017}, The Cancer Genome Atlas non-small cell lung cancer\footnote{http://www.cancer.gov/tcga} (TCGA-NSCLC), VisioMel\footnote{https://www.drivendata.org/competitions/148/visiomel-melanoma} and BReAst Carcinoma Subtyping (BRACS) \cite{brancati_bracs_2022} -- that: a) have an increasing clinical complexity, b) comprise both binary and multiple-class classification tasks and c) present different levels of class imbalance. \\
\textbf{Camelyon16} is a 2-class dataset for the detection of metastases in breast cancer. It includes 400 slides: 239 are normal tissue slides and 160 tumor slides. The official dataset is already split into training and test sets. For CAMx10, following \cite{li_dual-stream_2021}, we randomly split the official training set into training and validation sets at a 9:1 ratio. We tune all hyper-parameter values on the validation set. Instead, for CAMx20, we implement a 3-fold cross-validation on the training set.\\ 
\textbf{TCGA-NSCLC} is a 2-class, balanced lung cancer subtyping dataset with a total of 997 WSIs: 482 for Lung Squamous Cell Carcinoma (LUSC) and 515 for Lung Adenocarcinoma (LUAD). The WSIs are randomly divided into training, validation, and test sets with a ratio of 65:15:20.\\
\textbf{VisioMel} is used for the detection of the primary melanoma relapse. 
It is a 2-class, highly-imbalanced dataset that comprises 1342 cases: 213 positive cases and 1139 negative cases. As the test set is not yet released, we randomly split the official training set into training and validation sets at a 9:1 ratio. We report performances on the validation set.\\
\textbf{BRACS} is a 3-class imbalanced dataset. It contains 547 WSI for breast carcinoma subtyping out of which 265 are Benign Tumor, 89 are Atypical Tumor, and 193 are Malignant Tumor cases. We use the official split of 395, 65 and 87 samples for train, validation, and test sets, respectively. 

\begin{figure}
    \centering
    \includegraphics[width=0.8\linewidth]{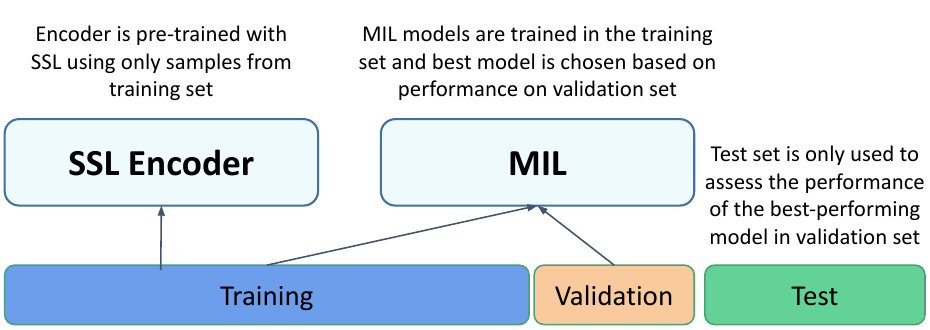}
    \caption{Visual explanation of the proposed validation pipeline. }
    \label{fig:sets}
\end{figure}

Our main evaluation metric is the AUC score which is resilient to class imbalance effects. We select the best-performing models on the validation set and report their AUC scores on the test sets, see Fig. \ref{fig:sets}.  Only for VisioMel, where the test set has not been released, we report the AUC scores on the validation set but, to limit over-optimistic results (and overfitting), we select the model from the last epoch, rather than the one with the best validation loss. 
\subsection{Pre-processing} We follow the CLAM's pre-processing pipeline \cite{lu_data-efficient_2021} for all datasets.  {Following prior literature, we cut all WSIs into 256x256 non-overlapping patches solely from foreground tissue regions at x10 magnification, the most commonly used resolution.} Only for Camelyon16, we extract patches both at x10 (\textit{CAMx10}) and x20 (\textit{CAMx20}) magnification for a more comprehensive analysis. This process resulted in a total of 4.6 million patches for CAMx20, 0.6 million patches for CAMx10, 2.7 million for TCGA-NSCLC, 1.4 million for BRACS and 2.7 million patches for VisioMel.  {We chose 256×256 over 224x224, as it results in fewer patches per slide, which helps reduce computational overhead.}
\subsection{Self-Supervised Learning Pre-Training Details} We include in our study ImageNet pre-training as a baseline and 6 different SSL methods: SimCLR \cite{chen_simple_2020}, MoCoV3 \cite{chen_empirical_2021}, MAE \cite{he_masked_2022}, DINO \cite{caron_emerging_2021}, BYOL \cite{grill_bootstrap_2020} and Barlow Twins \cite{zbontar_barlow_2021}. With the exception of MAE \cite{he_masked_2022} (compatible only with ViT), these methods can be used with arbitrary backbones. We experiment with four of the most used backbones in histopathological studies: \textit{ResNet18} (11.7M parameters), \textit{ResNet50} (25.5M parameters) \cite{he2016deep}, \textit{ViT-Tiny} (5.8M parameters) and \textit{ViT-Small} (22.2M parameters) 
 \cite{dosovitskiy_image_2021}. All SSL models are initialized using the Imagenet-1K weights. Please note that the foundation models CTransPath and UNI are based on two different architectures. In particular, UNI uses a ViT\_Large backbone, which has 307 millions parameters, and it is thus much larger than the backbones used in this work. All pre-trainings are done using the \textit{solo-learn} library from \cite{costa_solo-learn_2022} for 200 epochs. Regarding the SSL hyper-parameters, we keep the same values as in the original papers, except for the crop scale factor, which is reduced to range \{0.5, 1\}, to preserve the size of cells, as larger cells might be tumor cells. 
Additionally, for the learning rate (\textit{lr}) of each SSL method, as well as for the temperature parameter $\tau$ of SimCLR, we perform a grid-search, choosing as optimal value the one giving the best average AUC score across all tested MIL models in the validation set.
Following \cite{costa_solo-learn_2022}, we use LARS (respectively AdamW) for CNN (resp. ViT) backbones, and we use a warm-up phase of 10 epochs with a cosine annealing learning rate scheduler. 
Furthermore, we select the biggest \textit{batchsize} that fits in the GPU memory of our computational cluster  {as previous works consistently show that larger batch sizes yield better results. We choose the maximum batch size that fits in our GPU memory for each experiment.} \textit{ResNet18} and \textit{ViT-Tiny} backbones are trained on 4 NVIDIA V100 GPUs with \textit{batchsize} 256 per GPU, resulting in a global \textit{batchsize} of 4x256=1024. In order to keep the same \textit{batchsize} also for larger \textit{ResNet50} and \textit{ViT-Small} backbones, we used 8 NVIDIA V100 GPUs with a \textit{batchsize} of 128 for each, 8x128=1024. The hyperparameters for all the combinations of SSL methods, backbones, and datasets can be found in our Github {repository}\footnote{\url{https://github.com/mammadov7/wsi_classification_pipeline}} in the following directory:\textit{ pre-train/scripts/pretrain/custom}. In Table \ref{hyper}, as an example, we show the hyperparameters used for the Barlow Twins SSL method with ResNet18 backbone on the TCGA-NSCLC dataset. The config file path is \textit{(pre-train/scripts/pretrain/custom/tcga/barlow-res.yaml)} in our repository.
\begin{table*}[]
    \centering
    \caption{{ Self-Supervised Pre-Training Hyperparameters, where \textit{P} is probability} and \textit{S} is scale}
    \small 
    \resizebox{0.7\textwidth}{!}{
    \begin{tabular}{cc|cc}
    \toprule
    \multicolumn{2}{c}{\textbf{Training}} & \multicolumn{2}{c}{\textbf{Augmentations}}  \\
     \cmidrule(lr){1-4}  
     \textit{\textbf{Optimizer}} & LARS & \textit{\textbf{Crop}} & S = (0.5, 1.0) \\
     \textit{\textbf{Scale Loss ($\lambda$)}} & 0.1 & \textit{\textbf{ColorJitter}} & S = (0.4,0.4,0.2,0.1), P = 0.8 \\
     \textit{\textbf{Scheduler}} & Cosine Annealing & \textit{\textbf{Grayscale}} & P = 0.2 \\
     \textit{\textbf{Base Learning Rate}} & 0.3 & \textit{\textbf{GaussianBlur}} & S = (0.1, 2.0), P = 1.0 \\
     \textit{\textbf{Weight Decay}} & 1e-6 & \textit{\textbf{Solarization}} & P = 0.0 \\
     \textit{\textbf{Batch Size}} & 256*4, (4 GPUs) & \textit{\textbf{Horizontal Flip}} & P = 0.5 \\
    \cmidrule(lr){1-4}  
    \end{tabular}   }
    \label{hyper}
\end{table*}

\subsection{Pathology Adapted SSL methods details } 
 As proposed by \cite{kang_benchmarking_2023}, we apply pathology-specific augmentations, vertical flipping and stain augmentations, to three SSL methods (Barlow Twins, MOCOv3, BYOL)  to assess whether these adaptations can further enhance the results. However, we exclude multiple magnification levels due to their high computational cost. All the other training components and hyperparameters remain unchanged, as described in the previous subsection. Unfortunately, we could not include the \textit{SRCL} method, due to the lack of code in the source repository. For CluBYOL, we use the same hyper-parameters as in the original paper~\cite{wu_improving_2023} and recompute the statistics on the validation set.

\subsection{Foundation Models Details}
We perform no training or fine-tuning of the foundation models. Instead, we download the pre-trained weights from their original repositories and load them into the corresponding backbones, as detailed in Section 2.6. We then use these models directly to extract features from the patches, which we further use for MIL trainings.

\subsection{MIL Training Details} 
\begin{figure}[h]
    \centering
    \includegraphics[width=\textwidth]{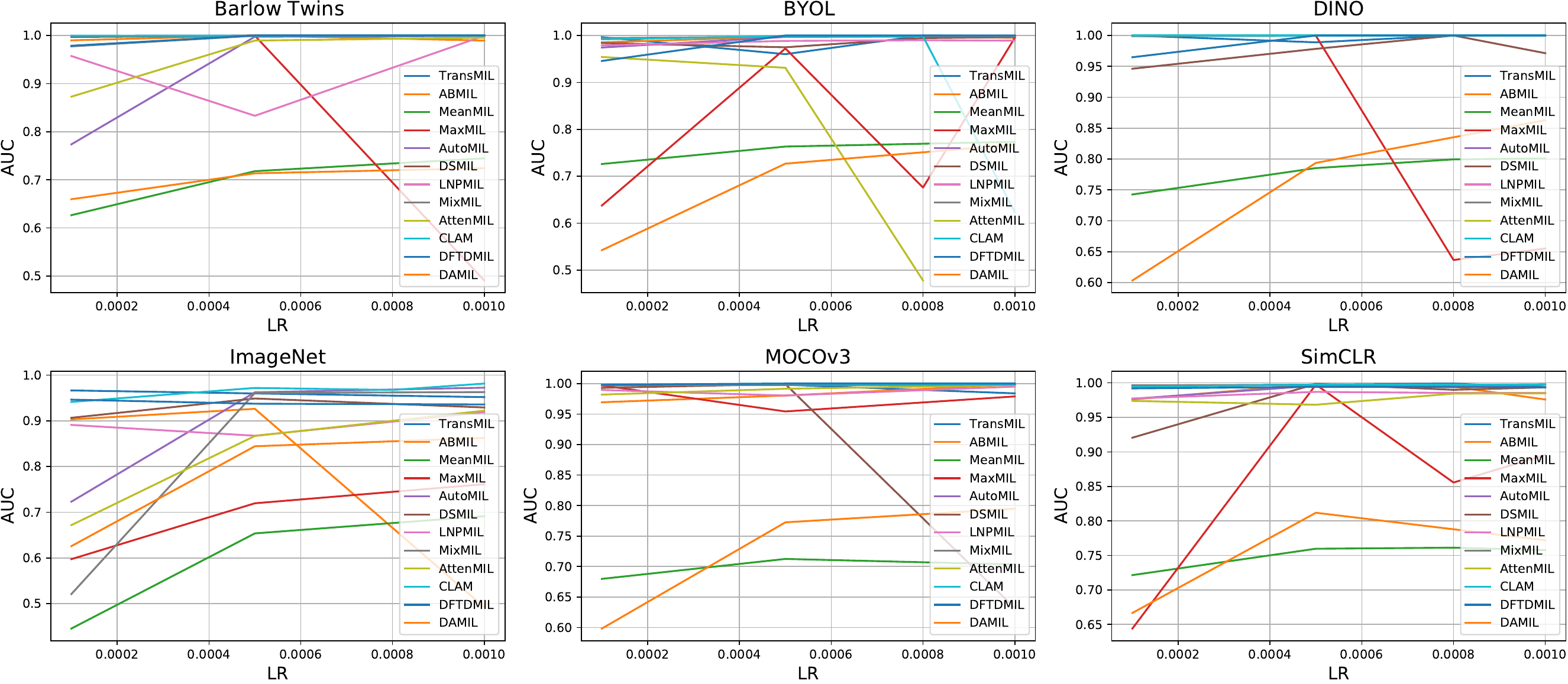} 
  { \caption{Grid Search of Learning Rate on Validation set of Camelyon16 dataset with 6 pre-training methods and 12 MILs with ResNet50 backbone } } 
  \label{fig:grid_lr_2}
\end{figure}

We adopt DSMIL’s code \cite{li_dual-stream_2021} as a base for our training and evaluation pipeline. Then, for each MIL method, we replace the model, optimizers and other training components based on their official codes. We trained each MIL model for 100 \textit{epochs} with a grid search for optimal \textit{learning\_rate}.  {Fig.} \ref{fig:grid_lr_2}  {illustrates the applied grid search process. The figure consists of six subplots, each corresponding to a different pre-training SSL method. In each subplot, the Validation AUC scores of 12 MIL methods are plotted on the Y-axis, while the learning rate is represented on the X-axis. For example, in the case of SimCLR, we select 0.0005 as the optimal learning rate for MaxMIL, as it achieves the highest validation score.} We use the cosine annealing scheduler, Adam optimizer with a weight decay of 0.00001 and a batch size equal to 1 slide (\textit{i.e.,} bag).  {For MIL, due to the varying number of patches per slide, the batch size is always 1. This is a standard practice in MIL literature and is followed in all related works. The bag size depends on the number of patches per slide. Since this varies across slides, it is not a fixed hyperparameter.} For more details about the other hyper-parameters please refer to the released code. 

\section{Results and Discussions}  
In this section, we share our results from all datasets
and discuss our findings based on quantitative and qualitative results.

\begin{table*}[]
    \caption{  Camelyon16 dataset at 10x. First and second-best AUC scores in bold and underlined respectively. The best average AUC scores (per MIL) in red. ImageNet-based results are not averaged.  {A Welch T-Test was applied to compare the best performing method (Ref.) with the others, where ($*$) indicates statistically significant differences ($p < 0.05$) and "N/S" denotes not significance. The \textit{DAMIL} method introduces additional complexity because the K-Means algorithm is applied independently on each slide.} }
    \centering
    \resizebox{\textwidth}{!}{
    \begin{tabular}{cccccccc|cccccc|c}
    \toprule
    \multirow{2}{*}{ \rotatebox{90}{\makecell{\textbf{Arch.}}}} &     
    \textbf{Pretrain.} &\textit{MeanMIL} & \textit{MaxMIL} & \textit{MixMIL} & \textit{AutoMIL} & \textit{LNPMIL} & \textit{AttenMIL} & \textit{ABMIL} & \textit{DSMIL} & \textit{CLAM} & \textit{TransMIL} &  {\textit{DTFDMIL}} &  {\textit{DAMIL}} &  {\textit{Avg.}} \\
    \cmidrule(lr){2-15}  
    & \textbf{\# Params.} &  (193)  & (193)  & (194)  & (194)  & (194)  & (386)  & (0.03M)  & (0.06M)  & (0.23M)  &  (2.25M) &  (0.18M) &  (0.02M) &  (-) \\
    \cmidrule(lr){1-15}  
    \multirow{8}{*}{\rotatebox{90}{\makecell{ViT-Tiny}}} 
    & \textit{ImageNet}         & 58.1 & 69.2 & 72.5 & 71.6 & 71.5 & 73.9 & 65.5 & 70.5 & 71.2 & 66.7 & 71.4 & 58.6 & 68.4$\pm$5.0 \\
    \cmidrule{2-15} 
    & Barlow T.                 & 58.5 & 74.1 & 79.7 & 77.9 & \textbf{83.8} & \underline{82.6} & 71.9 & 77.6 & 79.3 & 75.1 & 70.7 & 69.6 & 75.1$\pm$6.6 \\
    & MOCOv3                    & 51.8 & 75.2 & 75.9 & 70.7 & 80.7 & 74.3 & 72.7 & 76.6 & 73.8 & 72.0 & 80.2 & 71.4 & 72.9$\pm$7.1 \\
    & SimCLR                    & 61.0 & 76.0 & 74.2 & 77.4 & 73.8 & 75.4 & 76.9 & 72.8 & 77.5 & 72.7 & 76.4 & 63.2 & 73.1$\pm$5.2 \\
    & BYOL                      & 63.0 & 72.1 & 73.7 & 73.8 & \underline{82.6} & 77.4 & 74.4 & 74.5 & 69.2 & 78.5 & 74.2 & 52.3 & 72.1$\pm$7.5 \\
    & DINO                      & 64.5 & 77.1 & 76.8 & 74.6 & 75.3 & 72.8 & 73.7 & 81.6 & 73.0 & 73.8 & 48.2 & 63.7 & 71.3$\pm$8.4 \\
    & MAE                       & 50.6 & 57.4 & 59.5 & 54.2 & 61.9 & 66.9 & 72.1 & 67.4 & 64.7 & 65.2 & 71.2 & 56.6 & 62.3$\pm$6.5 \\
    \cmidrule(lr){2-15}  
    & \textbf{\textit{Avg.}}    & 58.2$\pm$5.3 & 72.0$\pm$6.7 & 73.3$\pm$6.5 & 71.4$\pm$8.1 & \red{\textbf{76.4}}$\pm$7.4 & 74.9$\pm$4.7 & 73.6$\pm$1.7 & \red{\underline{75.1}}$\pm$4.4 & 72.9$\pm$4.9 & 72.9$\pm$4.0 & 70.1$\pm$10.3 & 62.8$\pm$6.7 & 71.1$\pm$8.1 \\
    \cmidrule(lr){2-15}  
    &  {\textit{p-value}}          & * &   N/S &   N/S &   N/S & Ref. &  N/S &   N/S &   N/S &   N/S &   N/S &   N/S &   * & - \\
    \cmidrule(lr){1-15}  
     & \textbf{\# Params.} &  (385)  & (385)  & (386)  & (386)  & (386)  & (770)  & (0.05M)  & (0.2M)   & (0.33M)  &  (2.34M) &  (0.23M) &  (0.03M) &  (-)  \\
    \cmidrule(lr){2-15}  
    \multirow{8}{*}{\rotatebox{90}{\makecell{ViT-Small}}} 
    & \textit{ImageNet}         & 57.4 & 64.8 & 77.2 & 75.4 & 75.0 & 72.6 & 76.1 & 74.9 & 71.8 & 73.6 & 71.1 & 64.4 & 71.2$\pm$5.7 \\
    \cmidrule{2-15} 
    & Barlow T.                 & 50.4 & 87.6 & 85.3 & 81.2 & 84.3 & 85.4 & 78.3 & 77.3 & 84.1 & 84.6 & 85.2 & 64.3 & 79.0$\pm$10. \\
    & MOCOv3                    & 58.5 & 93.3 & \underline{93.5} & \textbf{94.8} & 93.4 & 91.5 & 92.5 & \underline{93.5} & 93.2 & 91.3 & 92.8 & 62.9 & 87.6$\pm$12. \\
    & SimCLR                    & 60.5 & 73.2 & 72.5 & 67.5 & 76.9 & 63.5 & 75.5 & 76.1 & 69.2 & 69.0 & 79.0 & 76.3 & 71.6$\pm$5.5 \\
    & BYOL                      & 62.9 & 85.9 & 89.0 & 83.3 & 90.8 & 80.1 & 84.4 & 86.4 & 86.8 & 84.8 & 85.7 & 69.3 & 82.4$\pm$7.9 \\
    & DINO                      & 63.2 & 82.7 & 84.1 & 84.4 & 80.1 & 84.4 & 85.8 & 81.8 & 75.8 & 82.2 & 78.7 & 63.8 & 78.9$\pm$7.4 \\
    & MAE                       & 50.2 & 70.1 & 72.3 & 68.4 & 60.6 & 74.5 & 79.0 & 77.3 & 73.8 & 78.4 & 70.5 & 54.8 & 69.2$\pm$8.9 \\
    \cmidrule(lr){2-15}  
    & \textbf{\textit{Avg.}}    & 57.6$\pm$5.4 & 82.1$\pm$8.1 & \red{\textbf{82.8}}$\pm$7.9 & 79.9$\pm$9.5 & 81.0$\pm$11. & 79.9$\pm$9.0 & \red{\underline{82.6}}$\pm$5.7 & 82.1$\pm$6.2 & 80.5$\pm$8.3 & 81.7$\pm$6.9 & 82.0$\pm$7.0 & 65.2$\pm$6.5 & 78.1$\pm$11. \\
    \cmidrule(lr){2-15}  
    &  {\textit{p-value}}          & * &   N/S & Ref. &  N/S &   N/S &   N/S &   N/S &   N/S &   N/S &   N/S &   N/S &   * & - \\
        \cmidrule(lr){1-15}  
    & \textbf{\# Params.}  &  (513)  & (513)  & (514)  & (514)  & (514)  & (1026) & (0.07M)  & (0.33M)  & (0.4M)   &  (2.41M)  &  (0.26M) &  (0.03M) &  (-) \\
        \cmidrule(lr){2-15}      
    \multirow{7}{*}{\rotatebox{90}{\makecell{ResNet18}}}
    & \textit{ImageNet}         & 56.2 & 66.8 & 73.4 & 70.5 & 67.8 & 72.4 & 72.9 & 60.1 & 62.5 & 70.2 & 73.1 & 64.3 & 67.5$\pm$5.4 \\
    \cmidrule{2-15} 
    & Barlow T.                 & 71.0 & 87.3 & 90.4 & 88.4 & 94.8 & 88.5 & 88.1 & 91.2 & 88.2 & 91.5 & 93.9 & 66.9 & 86.7$\pm$8.3 \\
    & MOCOv3                    & 72.3 & 87.6 & 88.5 & 87.9 & \textbf{96.0} & 84.9 & 87.0 & 88.7 & 82.6 & 90.2 & 89.2 & 65.0 & 85.0$\pm$8.1 \\
    & SimCLR                    & 72.6 & 89.4 & 90.9 & 90.0 & \underline{95.1} & 91.4 & 86.5 & 90.3 & 93.8 & 92.6 & 81.7 & 62.5 & 86.4$\pm$9.3 \\
    & BYOL                      & 60.7 & 73.6 & 74.7 & 77.4 & 74.4 & 75.5 & 74.1 & 73.8 & 75.2 & 75.1 & 78.5 & 53.1 & 72.2$\pm$7.1 \\
    & DINO                      & 69.2 & 89.0 & 90.8 & 90.3 & 92.7 & 90.8 & 88.3 & 88.6 & 84.4 & 91.9 & 92.0 & 72.2 & 86.7$\pm$7.5 \\
    \cmidrule(lr){2-15}  
    & \textbf{\textit{Avg.}}    & 69.2$\pm$4.4 & 85.4$\pm$5.9 & 87.1$\pm$6.2 & 86.8$\pm$4.8 & \red{\textbf{90.6}}$\pm$8.2 & 86.2$\pm$5.8 & 84.8$\pm$5.4 & 86.5$\pm$6.4 & 84.8$\pm$6.2 & \red{\underline{88.3}}$\pm$6.6 & 87.1$\pm$6.0 & 63.9$\pm$6.3 & 83.4$\pm$9.9 \\
    \cmidrule(lr){2-15}  
    &  {\textit{p-value}}          &  *  & N/S  & N/S  & N/S  & Ref. & N/S  & N/S  & N/S  & N/S  & N/S  & N/S  & * & - \\ 
    \cmidrule(lr){1-15}  
    \end{tabular} }
    \label{cam10x}
\end{table*}

\begin{table*}[]
    \centering
    \caption{{  Camelyon16 dataset at 20x. First and second-best AUC scores are in bold and underlined respectively. The best average AUC scores (per MIL) in red. ImageNet-based results are not averaged.  {A Welch T-Test was applied to compare the best performing method (Ref.) with the others, where ($*$) indicates statistically significant differences ($p < 0.05$) and "N/S" denotes not significance. The \textit{DAMIL} method introduces additional complexity because the K-Means algorithm is applied independently on each slide.} }}
    \resizebox{\textwidth}{!}{
    \begin{tabular}{ccccccc|cccccc|c}
    \toprule
    \multirow{2}{*}{ \rotatebox{90}{\makecell{\textbf{Arch.}}}} &     
    \textbf{Pretrain.} & \textit{MaxMIL} & \textit{MixMIL} & \textit{AutoMIL} & \textit{LNPMIL} & \textit{AttenMIL} & \textit{ABMIL} & \textit{DSMIL} & \textit{CLAM} & \textit{TransMIL} &  {\textit{DTFDMIL}} &  {\textit{DAMIL}} &  {\textit{Avg.}} \\ 
    \cmidrule(lr){2-14}  
     & \textbf{\# Params.} & (2049) & (2050) & (2050) & (2050) & (4098) & (0.26M)  & (4.46M)  & (1.18M)  &  (3.2M)  &  (0.66M)  &  (0.14M)  &  (-)  \\
    \cmidrule(lr){1-14}  
    \multirow{7}{*}{\rotatebox{90}{\makecell{ResNet50}}}
   & \textit{ImageNet}          & 64.4 & 83.2 & 77.7 & 77.1 & 72.0 & 77.2 & 80.7 & 86.3 & 78.9 & 77.4 & 63.4 & 76.2$\pm$6.8 \\ 
   \cmidrule{2-14}   
    & Barlow T.                 & 98.3 & 98.5 & 96.1 & 97.2 & 96.4 & 98.4 & 98.5 & 98.1 & 98.3 & 94.3 & 65.1 & 94.5$\pm$9.4  \\
    & MOCOv3                    & 96.7 & 97.4 & 97.0 & 94.3 & 96.8 & 97.2 & 96.7 & 97.5 & 95.8 & 97.1 & 66.7 & 93.9$\pm$8.7  \\
    & SimCLR                    & 81.4 & 92.3 & 92.4 & 84.5 & 91.2 & 89.9 & 88.0 & 92.4 & 92.0 & 94.8 & 62.2 & 87.4$\pm$8.8  \\
    & BYOL                      & 96.4 & 97.1 & 96.0 & 90.9 & 94.4 & 95.9 & 94.5 & 97.9 & 97.9 & 94.9 & 61.4 & 92.5$\pm$10.  \\
    & DINO                      & 98.4 & \textbf{99.1} & 98.3 & 94.7 & 98.0 & 97.9 & 93.9 & 98.0 & 98.5 & \underline{98.8} & 64.4 & 94.6$\pm$9.7  \\
    \cmidrule(lr){2-14}  
    & \textbf{\textit{Avg.}}    & 94.2$\pm$6.5 & \red{\textbf{96.9}}$\pm$2.4 & 96.0$\pm$2.0 & 92.3$\pm$4.4 & 95.4$\pm$2.4 & 95.9$\pm$3.1 & 94.3$\pm$3.6 & \red{\underline{96.8}}$\pm$2.2 & 96.5$\pm$2.4 & 96.0$\pm$1.7 & 64.0$\pm$1.9 & 92.6$\pm$9.7 \\
    \cmidrule(lr){2-14}  
    &  {\textit{p-value}}          & N/S & Ref. &   N/S &   N/S &   N/S &   N/S &   N/S &   N/S &   N/S &   N/S &   * & - \\
     \cmidrule(lr){1-14}  
    \end{tabular}} 
    \label{cam20x}
\end{table*}

\begin{table*}[]
    \centering
    \caption{{  Foundation models on Camelyon16 dataset at 10x. First and second-best AUC scores in bold and underlined respectively. Best average AUC score in red. }}
    \small 
    \resizebox{0.9\textwidth}{!}{
    \begin{tabular}{ccccccc|cccc}
    \toprule
    \multirow{2}{*}{\textbf{Pretrain}} & \multicolumn{10}{c}{\textbf{Camelyon16}}  \\
     \cmidrule(lr){2-11}  
    & \textit{MeanMIL} & \textit{MaxMIL} & \textit{MixMIL} & \textit{AutoMIL} & \textit{LNPMIL} & \textit{AttenMIL} & \textit{ABMIL} & \textit{DSMIL} & \textit{CLAM} & \textit{TransMIL} \\
    \cmidrule(lr){1-11}  
     SRCL({ CTransPath})           & 57.3 & 91.1 & 89.8 & 92.4 & 94.7 & 91.9 & 89.2 & 91.3 & 88.0 & 94.7 \\
     PathAugFM({ ViT\_Small})    & 60.2 & 91.5 & 95.3 & 95.8 & 90.9 & 93.6 & 94.4 & 91.7 & 94.7 & 91.4 \\
     DINOv2({ ViT\_Small})         & 60.0 & 46.3 & 76.4 & 80.1 & 70.3 & 55.6 & 76.0 & 49.9 & 79.6 & 77.1 \\
     UNI({ ViT\_Large})            & 64.5 & 97.7 & 96.8 & 96.8 & \underline{98.8} & 96.7 & 97.4 & 95.7 & \textbf{99.1} & 98.3 \\
    \cmidrule(lr){1-11}       
    \textbf{\textit{Avg.}}         & 60.5$\pm$2.6 & 81.6$\pm$21. & 89.6$\pm$8.0 & \red{\textbf{91.3}}$\pm$6.7 & 88.7$\pm$11. & 84.4$\pm$17. & 89.2$\pm$8.2 & 82.2$\pm$19. & \red{\underline{90.4}}$\pm$7.4 & 90.4$\pm$8.0 \\
     \cmidrule(lr){1-11}  
    \end{tabular}}
    \label{fm}
\end{table*}

\begin{table*}
    \centering
    \caption{{AUC scores of different MIL methods with Pathology Adapted SSL methods on Camelyon16 dataset at 10x magnification, \textit{{\scriptsize -path}} suffix indicates methods with adapted augmentations. The red/green numbers show if adapted augmentations downgraded/improved results: $method_{path} - method$.}}
    \small 
    \resizebox{0.9\textwidth}{!}{
    \begin{tabular}{ccccccccccc|c}
    \toprule
    \multirow{2}{*}{\textbf{Pretrain}} & \multicolumn{11}{c}{\textbf{Camelyon16}}  \\
     \cmidrule(lr){2-12}  
    & \textit{MeanMIL} & \textit{MaxMIL} & \textit{MixMIL} & \textit{AutoMIL} & \textit{LNPMIL} & \textit{AttenMIL} & \textit{ABMIL} & \textit{DSMIL} & \textit{CLAM} & \textit{TransMIL} & \textbf{Avg.} \\
    \cmidrule(lr){1-12}  
    Barlow         & 71.0 & 89.3 & 90.4 & 88.4 & 94.8 & 88.5 & 88.1 & 91.2 & 88.2 & 91.5 & 87.9 \\
    $Barlow_{path}$  & 69.4\red{-1.6} & 93.5\green{+6.2} & 93.1\green{+2.7} & 94.6\green{+6.2} & 94.9\green{+0.1} & 92.3\green{+3.8} & 87.4\red{-0.7} & 88.9\red{-2.3} & 90.6\green{+2.4} & 88.8\red{-2.7} & 89.4\green{+1.5}\\ 
    MOCOv3         & 72.3 & 87.6 & 88.5 & 87.9 & 96.0 & 84.9 & 87.0 & 88.7 & 82.6 & 90.2 & 86.6\\
    $MOCOv3_{path}$  & 71.2\red{-1.1} & 92.3\green{+4.7} & 92.4\green{+3.9} & 93.5\green{+5.6} & 91.8\red{-4.2} & 92.6\green{+7.7} & 79.2\red{-7.8} & 85.3\red{-3.4} & 86.5\green{+3.9} & 93.4\green{+3.2} & 87.8\green{+1.2}\\
    BYOL           & 60.7 & 73.6 & 74.7 & 77.4 & 74.4 & 75.5 & 74.1 & 73.8 & 75.2 & 75.1 & 73.5\\
    $BYOL_{path}$   & 64.9\green{+4.2} & 75.1\green{+1.5} & 78.4\green{+3.7} & 80.2\green{+2.8} & 71.0\red{-3.4} & 82.0\green{+6.5} & 81.1\green{+7.0} & 80.3\green{+6.5} & 80.6\green{+5.4} & 74.8\red{-0.3} & 76.8\green{+3.3}\\
    CluBYOL        & 58.4 & 89.7 & 89.4 & 90.1 & 91.4 & 70.4 & 74.9 & 89.3 & 85.2 & 89.0 & 82.1\\
    $CluBYOL_{path}$ & 59.8\green{+1.4} & 88.8\red{-0.9} & 88.0\red{-1.4} & 89.0\red{-1.1} & 92.7\green{+1.3} & 86.8\green{+16.} & 83.4\green{+8.5} & 80.6\red{-8.7} & 87.3\green{+2.1} & 82.7\red{-6.3} & 83.9\green{+1.8}\\
     \cmidrule(lr){1-12}  
    \end{tabular}}
    \label{path}
\end{table*}

\begin{table*}[]
        \centering
    \caption{{Effect on the final AUC score of the \textit{ImageNet} initialization using ResNet18 as backbone. We compare three SSL techniques with all MIL methods on the Camelyon16 dataset at 10x magnification. Each result ( Average{\tiny std} ) is the average and standard deviation of three different trainings. \textit{IN} suffix stands for \textit{ImageNet} initialization, no suffix indicates random initialization and $\Delta = method \,IN - method$.
    }}
        \resizebox{0.9\textwidth}{!}{
        \begin{tabular}{cccccccccccc|c}
        \toprule
        
        \multirow{2}{*}{ \rotatebox{90}{\makecell{\textbf{Arch.}}}} &     
        \textbf{Pretrain.} &\textit{MeanMIL} & \textit{MaxMIL} & \textit{MixMIL} & \textit{AutoMIL} & \textit{LNPMIL} & \textit{AttenMIL} & \textit{ABMIL} & \textit{DSMIL} & \textit{CLAM} & \textit{TransMIL} & \textbf{Avg.} \\
        
     \cmidrule(lr){2-13}  
    & \textbf{Params.}  & (513)  & (513)  & (514)  & (514)  & (514)  & (1026) & (0.07M)  & (0.33M)  & (0.4M)   &  (2.41M) & (-) \\
    \cmidrule(lr){1-13}  
    \multirow{7}{*}{\rotatebox{90}{\makecell{ResNet18}}}
    & Barlow T.                & 71.8{\scriptsize $\pm$1.5} & 89.3{\scriptsize $\pm$0.4} & 88.8{\scriptsize $\pm$1.5} & 90.4{\scriptsize $\pm$0.9} & 95.1{\scriptsize $\pm$0.3} & 89.8{\scriptsize $\pm$0.4} & 86.3{\scriptsize $\pm$1.4} & 89.6{\scriptsize $\pm$0.4} & 89.2{\scriptsize $\pm$2.1} & 89.6{\scriptsize $\pm$1.4} & 87.7 \\
    & Barlow T.\textit{IN}     & 71.0{\scriptsize $\pm$1.3} & 87.3{\scriptsize $\pm$1.9} & 90.4{\scriptsize $\pm$1.3} & 88.4{\scriptsize $\pm$2.0} & 94.8{\scriptsize $\pm$0.1} & 88.5{\scriptsize $\pm$1.9} & 88.1{\scriptsize $\pm$1.6} & 91.2{\scriptsize $\pm$1.8} & 88.2{\scriptsize $\pm$1.4} & 91.5{\scriptsize $\pm$1.2} & 87.9 \\
    & Barlow T.\textit{IN} - Barlow T. & \red{-0.8} & \red{-2.0} &  \green{+1.6} & \red{-2.0} & \red{-0.3} & \red{-1.3} &  \green{+1.8} &  \green{+1.6} & \red{-1.0} &  \green{+1.9} & \green{+0.1} \\
    & SimCLR                   & 72.6{\scriptsize $\pm$0.2} & 89.4{\scriptsize $\pm$1.3} & 90.6{\scriptsize $\pm$1.8} & 92.1{\scriptsize $\pm$2.3} & 95.4{\scriptsize $\pm$0.5} & 91.9{\scriptsize $\pm$0.7} & 88.9{\scriptsize $\pm$2.3} & 91.2{\scriptsize $\pm$0.3} & 91.5{\scriptsize $\pm$1.6} & 90.2{\scriptsize $\pm$2.1} & 89.4 \\
    & SimCLR \textit{IN}        & 72.6{\scriptsize $\pm$0.4} & 89.4{\scriptsize $\pm$1.4} & 90.9{\scriptsize $\pm$2.2} & 90.0{\scriptsize $\pm$0.5} & 95.1{\scriptsize $\pm$0.5} & 91.4{\scriptsize $\pm$1.8} & 86.5{\scriptsize $\pm$1.7} & 90.3{\scriptsize $\pm$2.2} & 93.8{\scriptsize $\pm$0.2} & 92.6{\scriptsize $\pm$1.5} & 89.3 \\
    & SimCLR \textit{IN} - SimCLR & 0.0 &  0.0 &  \green{+0.3} & \red{-2.1} & \red{-0.3} & \red{-0.5} & \red{-2.4} & \red{-0.9} &  \green{+2.3} &  \green{+2.4} & \red{-0.1} \\
    & DINO                     & 69.7{\scriptsize $\pm$1.0} & 89.1{\scriptsize $\pm$1.7} & 91.9{\scriptsize $\pm$0.5} & 91.1{\scriptsize $\pm$0.8} & 94.6{\scriptsize $\pm$0.8} & 90.2{\scriptsize $\pm$0.9} & 87.7{\scriptsize $\pm$1.3} & 88.6{\scriptsize $\pm$1.1} & 83.8{\scriptsize $\pm$0.8} & 92.9{\scriptsize $\pm$2.1} & 88.0 \\
    & DINO \textit{IN}          & 69.2{\scriptsize $\pm$0.9} & 89.0{\scriptsize $\pm$2.3} & 90.8{\scriptsize $\pm$0.6} & 90.3{\scriptsize $\pm$0.6} & 92.7{\scriptsize $\pm$1.2} & 90.8{\scriptsize $\pm$2.0} & 88.3{\scriptsize $\pm$1.4} & 88.6{\scriptsize $\pm$1.0} & 84.4{\scriptsize $\pm$2.0} & 91.9{\scriptsize $\pm$1.6} & 87.6 \\
    & DINO \textit{IN} - DINO   & \red{-0.5} & \red{-0.1} & \red{-1.1} & \red{-0.8} & \red{-1.9} &  \green{+0.6} &  \green{+0.6} &  0.0 &  \green{+0.6} & \red{-1.0} & \red{-0.4} \\ 

        \cmidrule(lr){1-13} 
        \end{tabular}
        } 
        \label{IN}
    \end{table*}

\begin{figure}[h]
    \centering
    \includegraphics[width=\textwidth]{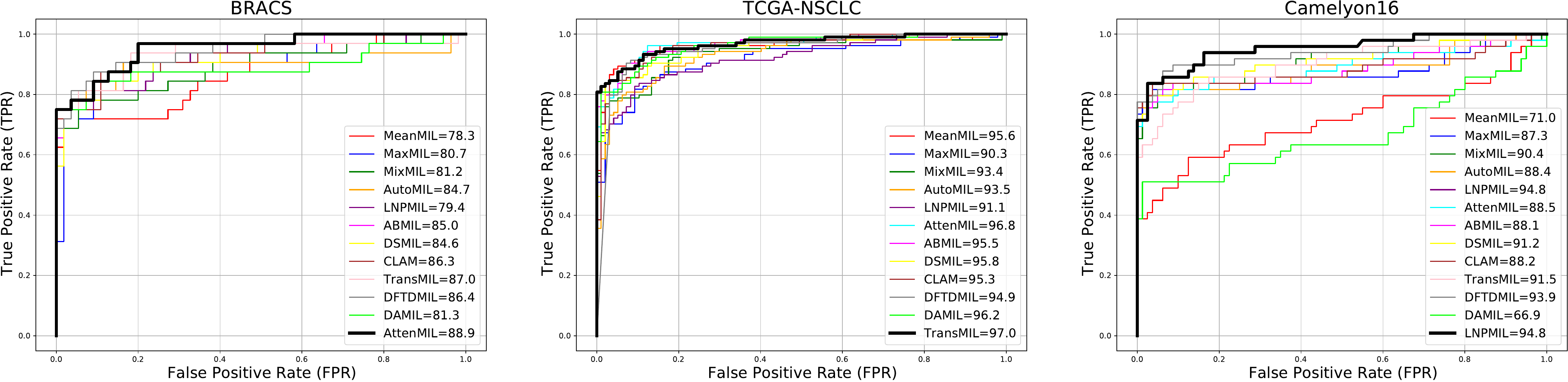} 
 {  \caption{ ROC curves for 12 MIL methods, features extracted by ResNet18 backbone pretrained with Barlow Twins. The curves are plotted across three datasets—BRACS, TCGA-NSCLC, and Camelyon16. The best-performing method on each dataset is highlighted in a black curve with a larger line width. }}
  \label{fig:rocs}
\end{figure}

\subsection{Camelyon16}
Tables \ref{cam10x}, \ref{cam20x},  \ref{fm}, \ref{path} present the classification results for Camelyon16 using various backbones, SSL, and MIL methods.\\
\noindent \textbf{Self-supervised pre-training. } Table \ref{cam10x} presents results for ViT-Tiny, ViT-Small, and ResNet18 backbones using data at x10 magnification, while Table \ref{cam20x} presents results for ResNet50 using data at x20 magnification. Across all setups, instance-based MIL methods are on par, or slightly better, than embedding-based MIL methods (on average over all SSL methods), despite having fewer parameters by several orders of magnitude. The best-performing combinations of SSL and MIL is based on instance-based MIL. Notably, the proposed MixMIL method, with a ResNet50 backbone pretrained with DINO on Camelyon16 at x20 magnification, achieves a new SOTA result with a 99.1 AUC score. \\
\noindent \textbf{Foundation models. } Table \ref{fm} reports MIL methods' performance when using foundation models as feature extractors, including the general-purpose DINOv2 model \cite{oquab_dinov2_2023} and in-domain CTransPath \cite{wang_transformer-based_2022}, PathAugFM~\cite{kang_benchmarking_2023} and UNI~\cite{chen2024uni}. 
We observe a similar trend between instance-based and embedding-based MIL methods, as with the previous SSL pre-trainings. We can also notice that the results of DINOv2 are significantly lower compared to the other models, which may be related to the domain gap between natural and pathological images.  \\
\noindent \textbf{Pathology-adapted SSL. } Table \ref{path} reports the effect of the pathology-adapted SSL methods on the performance of the MIL techniques. We compare three SSL methods (Barlow Twins, MOCOv3, BYOL), using the standard augmentations for natural images, as proposed in the original articles, and their histopathological counterparts, denoted with the \textit{-path} suffix, where we use the pathology-specific augmentations proposed in \cite{kang_benchmarking_2023}: vertical flipping and stain augmentation. For all methods, we use the ResNet18 backbone. We can see that results can vary quite a lot between standard and histopathological augmentations, going from a difference of -0.9 to a one of +16. However, on average, pathology-adapted augmentations improve performance by 1.2 to 3.3 AUC points across all MIL approaches. Additionally, CluBYOL, the pathology-adapted version of the BYOL framework, also performs better (+ 1.8 AUC points) when using pathology-specific augmentations.\\

\begin{figure}[h]
    \centering
    \includegraphics[width=\textwidth]{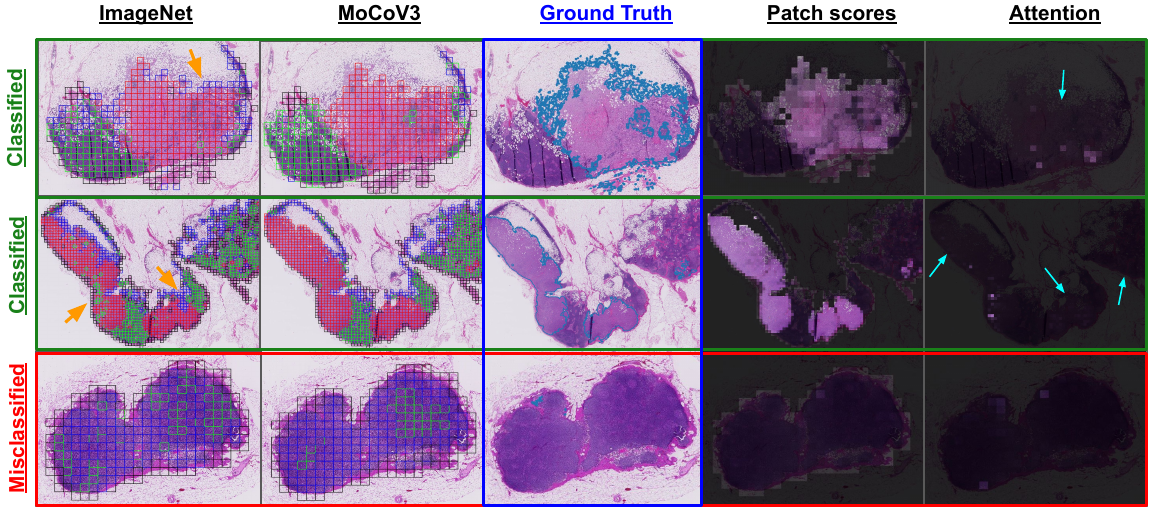} 
  \caption{\textit{Middle}: the $3^{rd}$ column presents the ground truth manual segmentation of the tumor areas from three samples of the Camelyon16 dataset. \textit{Left}: first two columns show a K-means clustering of the patch embeddings obtained with an ImageNet ($1^{st}$ column) and a MoCoV3 ($2^{nd}$ column) pre-training using ViT-Small as backbone. Orange arrows highlight the areas correctly clustered by MoCoV3 as homogeneous tumor areas but not by imageNet. \textit{Right}: the last two columns present the patch scores (\textit{i.e.,} $h(f(x))$) used in MaxMIL (max score is highlighted with a red square) and the Attention scores (\textit{i.e.,} $U(f_k,f_m)$) of DSMIL, respectively. Blue arrows indicate the areas that have been correctly identified as important by the instance-based MIL and not by the embedding-based one. The first two rows present correctly classified examples while the last row shows a sample that has been misclassified by all MIL methods. Both the patch and attention scores do not highlight the tumor area, which is also not identified in the two clustering of the embeddings. This is probably due to the fact that the tumor area is very small, with subtle changes. Future, well adapted SSL methods should be conceived for this kind of complex pathological data.}
  \label{fig:qualitative}
\end{figure}

\begin{figure}[!htbp]
    \centering
    \includegraphics[width=0.85\textwidth]{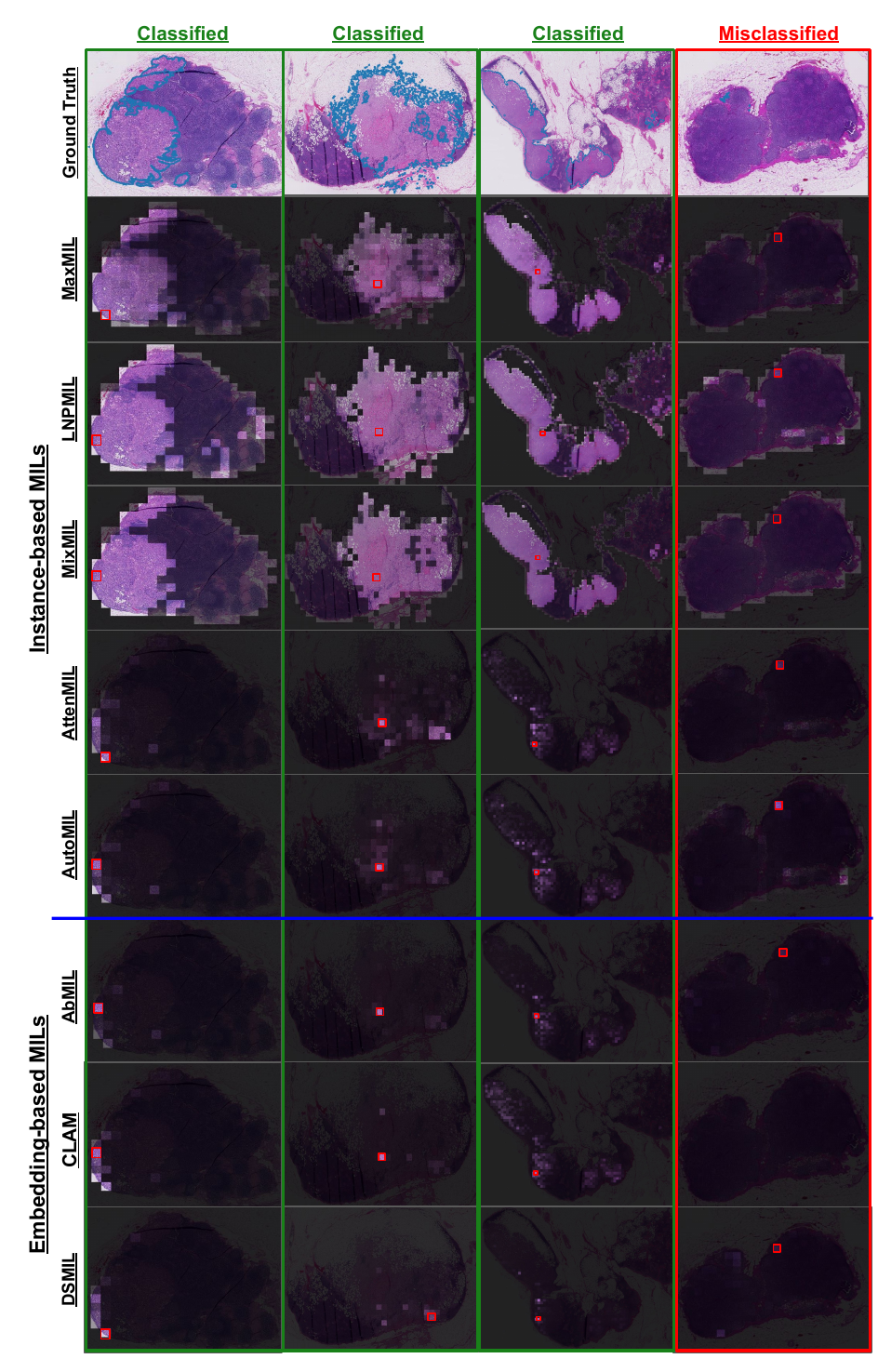}
    \vspace{-0.3cm}
    \caption{Qualitative results for the Camelyon16 dataset. Each column is one case from the dataset, green text means correct classification by all MIL models, while red indicates misclassification by all MIL models. Rows represent different MIL methods, with instance-based MILs above the blue line and embedding-based MILs below it. The red squares highlight patches with the highest scores, indicating that they have the maximum contribution to the final prediction. For MaxMIL, LNPMIL, and MixMIL scores are predictions of the instance-level classifier ($h(f(x_i))$ in section 3). For the AutoMIL, scores are based on $\frac{exp(\alpha \cdot y_k)}{ \sum_{j=1}^{K} exp(\alpha \cdot y_j)}$ . For the remaining MIL methods, we used their attention scores.
    }
    \label{fig:qualitative2}
\end{figure}

\noindent \textbf{Influence of ImageNet initialization. } Table~\ref{IN} presents the influence of the ImageNet initialization on the performance of three SSL techniques (Barlow Twins, SimCLR and DINO) combined with all MIL methods. We use as backbone the ResNet18 model. The \textit{-IN} suffix indicates an ImageNet initialization, while no suffix means that the model has been randomly initialized. We observe that, on average, the results do not indicate a significant impact from the ImageNet initialization (+0.1, -0.1 and -0.4 AUC points, respectively). This could be explained by the fact that training for 200 epochs is sufficient to effectively learn the representations of the different tissue types in the dataset, independently from the initialization. To evaluate the robustness of our results, we compute three different trainings per SSL/MIL combination and present both the average and standard deviation. Besides the random initialization of the SSL methods, the other important sources of variability come from the random initialization of the MIL methods and the random order of the batches.
Standard deviations are generally low, but they can go up to 2 AUC points. Nonetheless, even considering the estimated standard deviations, our conclusions and findings do not vary.\\
\noindent \textbf{Qualitative results. } In Fig.~\ref{fig:qualitative}, we present qualitative results from three different slides, two correctly classified and one wrongly classified by all MIL methods. On the left, we show how MoCoV3 (Vit-Small) produces better embeddings than ImageNet pre-training, which can be leveraged by simple instance-based MIL methods to classify the slides. Indeed, on the right, we can see that the patch scores well delineate the tumor area. However, this is not the case for the attention scores of DSMIL, which can not identify as important the entire tumor area, and are thus less interpretable than the patch scores. In Fig.~\ref{fig:qualitative2}, we present more qualitative results using all MIL models and MoCoV3 (Vit-Small) as SSL method. Similar conclusions as for Fig.~\ref{fig:qualitative} can be drawn when comparing instance-based and embedding-based MIL methods.

\begin{figure}[h]
\centering
\includegraphics[width=\textwidth]{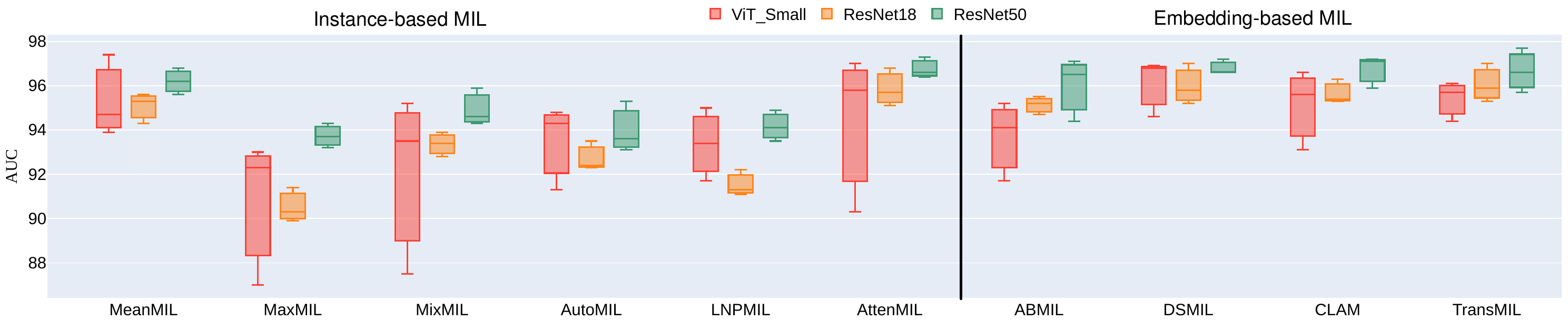}
  \caption{
  The Y-axis represents the AUC scores of different MIL methods for three backbones (ViT\_Small, ResNet18 and ResNet50) on the \textbf{TCGA-NSCLC} dataset at a resolution 10x. For each backbone and MIL method, we show the box-plot of 3 self-supervised pre-trainings: DINO, MOCO-V3 and Barlow Twins.
}
  \label{fig:tcga}
\end{figure}

\subsection{TCGA-NSCLC}
Table~\ref{tcga} and Fig.~\ref{fig:tcga} present classification results on the TCGA-NSCLC dataset. The performance of instance-based MILs is on par with embedding-based methods across all backbones. The best performing combinations of SSL and MIL are evenly spread across instance-based and embedding-based MILs. Note that embedding-based methods have 100 to 1000 times more parameters. 

\subsection{BRACS}
\begin{figure}[h]
\centering
\includegraphics[width=\textwidth]{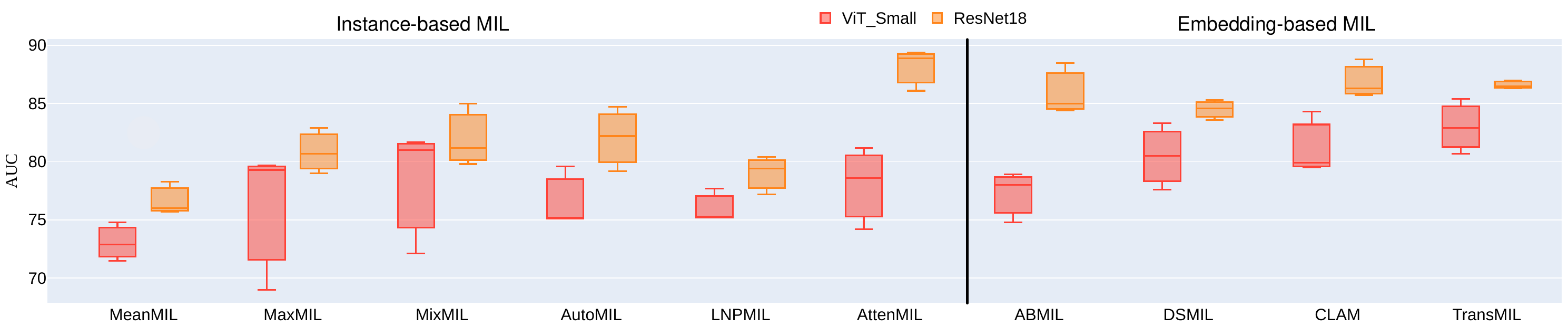}
  \caption{
  The Y-axis represents the AUC scores of different MIL methods for two backbones (ViT\_Small and ResNet18) on the \textbf{BRACS} dataset at a resolution 10x. For each backbone and MIL method, we show the box-plot of 3 self-supervised pre-trainings: DINO, MOCO-V3 and Barlow Twins.
}
  \label{fig:bracs}
\end{figure}

Table \ref{bracs} and Fig.~\ref{fig:bracs} present results for the multi-class classification task on the BRACS dataset. Reported AUC scores are averages over the 3 class AUC scores. Instance and embedding-based MIL methods reach similar AUCs, both when looking at the best combinations of MIL and SSL, and on average over all SSL  methods. Furthermore, the AttenMIL method with the ResNet18 backbone pretrained with DINO achieves a new SOTA result with a 89.4 AUC score. Finally, the lower overall scores on this dataset, with respect to previous datasets, can be related to the limited number of Atypical Tumor cases. 

\subsection{VisioMel}
\begin{figure}[h]
\centering
\includegraphics[width=\textwidth]{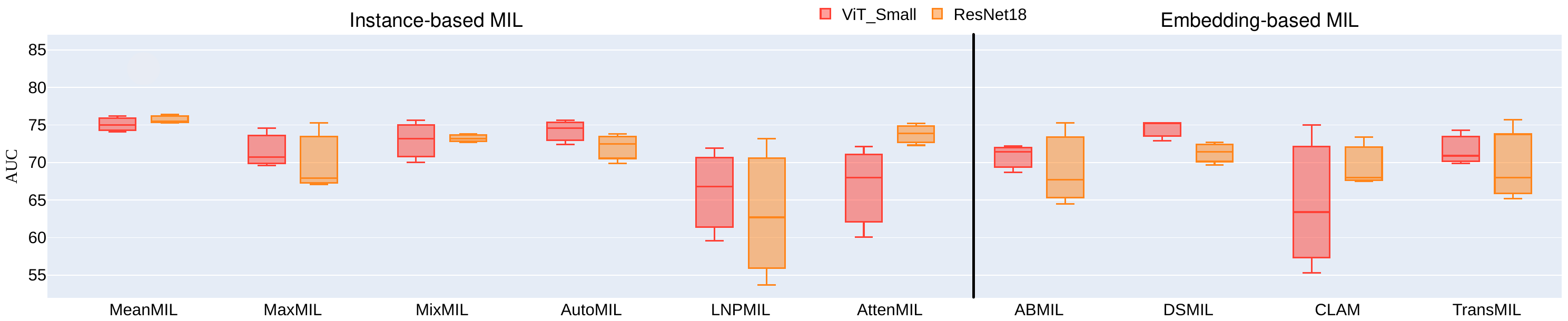}
  \caption{The Y-axis represents the AUC scores of different MIL methods for two backbones (ViT\_Small and ResNet18) on the \textbf{VisioMel} dataset at a resolution 10x. For each backbone and MIL method, we show the box-plot of 3 self-supervised pre-trainings: DINO, MOCO-V3 and Barlow Twins.
}
  \label{fig:mel}
\end{figure}

In Table \ref{mel} and Fig.~\ref{fig:mel}, we show the results on the VisioMel dataset. This is the most challenging clinical task, as indicated by the lowest overall AUC scores. Nevertheless, even here instance-based and embedding-based MIL methods perform similarly both in terms of best and averaged AUC scores. 

\subsection{Number of Epochs for Self-Supervised Pre-Training} 
\begin{figure}[h]
\centering
\includegraphics[width=0.8\textwidth]{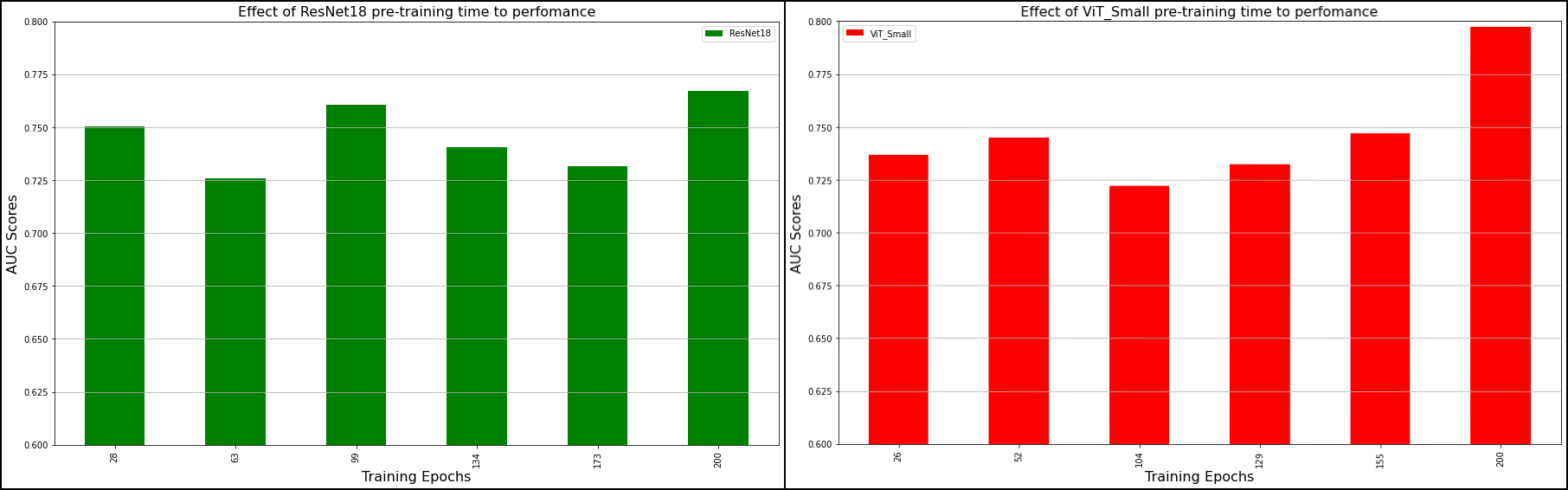}
  \caption{Effect of the number of epochs of SSL pre-training on the performance of a DSMIL model. On the left we use SimCLR with ResNet18 and on the right DINO with ViT\_Small.}
  \label{fig:time}
\end{figure}

In Fig. \ref{fig:time}, we illustrate the impact of the number of epochs of the SSL method on the performance of a MIL model. The x-axis denotes the number of epochs for the SSL pre-training, while the y-axis represents the average AUC scores of all MIL methods. Here, we show the results using SimCLR for ResNet18 and DINO for ViT\_Small  {on the Camelyon16 dataset}. In both examples, we show the AUC score obtained with the DSMIL method. Based on these experiments, we chose to train all SSL methods for 200 epochs, as extended pre-training enhances final performance and stabilizes the backbones, particularly for ViT ones.

\begin{figure}[h]
\centering
\includegraphics[width=0.6\textwidth]{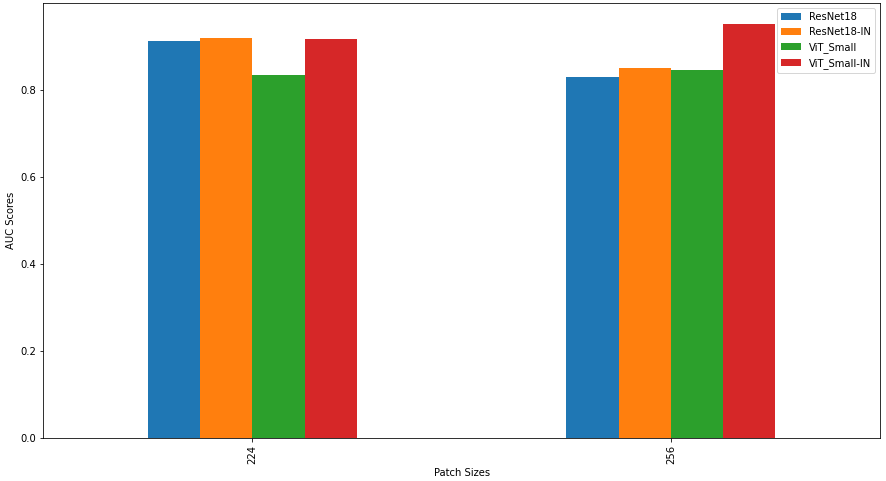}
  \caption{Effect of the patch size used in the SSL pre-training on the performance of a DSMIL method. We used SimCLR with ResNet18 and DINO with ViT\_Small. '-IN' means pre-trained on ImageNet}
  \label{fig:patch}
\end{figure}

\subsection{Patch size}
 In Fig.\ref{fig:patch}, we study the impact of the patch size used during SSL pre-training on the performance of a MIL method. We tested the two most used sizes in the literature: 224x224 and 256x256. As backbone, we used ResNet18 and Vit\_Small pre-trained (or not) on ImageNet ('-IN' means pre-trained on ImageNet). We show the results using SimCLR for ResNet18 and DINO for Vit\_Small  {on the Camelyon16 dataset}. For both patch sizes, we show the AUC score obtained with the DSMIL method.\\ 
We can notice that results are comparable using the two patch sizes. For computational reasons, we chose to train all SSL methods using a patch size of 256x256 (i.e., less patches and thus less computational time).

\begin{table*}[]
    \caption{  TCGA-NSCLC dataset at 10x. First and second-best AUC scores are in bold and underlined respectively. The best average AUC score (per MIL) is in red. ImageNet-based results are not averaged.  {A Welch T-Test was applied to compare the best performing method (Ref.) with the others, where ($*$) indicates statistically significant differences ($p < 0.05$) and "N/S" denotes not significance. The \textit{DAMIL} method introduces additional complexity because the K-Means algorithm is applied independently on each slide.}}
    \centering
    \resizebox{\textwidth}{!}{
    \begin{tabular}{cccccccc|cccccc|c}
    \toprule    
    \multirow{2}{*}{ \rotatebox{90}{\makecell{\textbf{Arch.}}}} &     
    \textbf{Pretrain.} &\textit{MeanMIL} & \textit{MaxMIL} & \textit{MixMIL} & \textit{AutoMIL} & \textit{LNPMIL} & \textit{AttenMIL} & \textit{ABMIL} & \textit{DSMIL} & \textit{CLAM} & \textit{TransMIL}  &  {\textit{DTFDMIL}} &  {\textit{DAMIL}} &  {\textit{Avg.}} \\   
    \cmidrule(lr){2-15}  
     & \textbf{\# Params.} &  (385)  & (385)  & (386)  & (386)  & (386)  & (770)  & (0.05M)  & (0.2M)   & (0.33M)  &  (2.34M) &  (0.23M) &  (0.03M) &  ( - ) \\
    \cmidrule(lr){1-15}  
    \multirow{8}{*}{\rotatebox{90}{\makecell{ViT-Small}}} 
    & \textit{ImageNet}         & 89.2 & 80.7 & 84.8 & 83.0 & 86.3 & 91.2 & 88.3 & 87.7 & 89.2 & 89.9 & 88.9 & 91.8 & 87.6$\pm$3.2 \\
    \cmidrule{2-15}
    & Barlow T.                 & 93.9 & 86.0 & 87.5 & 91.3 & 93.4 & 90.3 & 91.7 & 94.6 & 93.1 & 94.4 & 92.3 & 91.1 & 91.6$\pm$2.6 \\
    & MOCOv3                    & 94.7 & 92.3 & 93.5 & 94.3 & 95.0 & 95.8 & 95.2 & 96.9 & 95.6 & 95.7 & 95.8 & 96.7 & 95.1$\pm$1.2 \\
    & SimCLR                    & 71.2 & 65.3 & 64.4 & 72.8 & 67.5 & 64.0 & 69.2 & 76.2 & 65.6 & 61.7 & 67.2 & 70.1 & 67.9$\pm$4.0 \\
    & BYOL                      & 79.9 & 69.0 & 68.4 & 71.7 & 70.2 & 67.4 & 73.2 & 76.4 & 72.4 & 68.5 & 72.4 & 71.9 & 71.8$\pm$3.4 \\
    & DINO                      & \textbf{97.4} & 93.0 & 95.2 & 94.8 & 91.7 & \underline{97.0} & 94.1 & 96.8 & 96.6 & 96.1 & \textbf{97.4} & 96.8 & 95.6$\pm$1.8 \\
    & MAE                       & 81.3 & 79.3 & 77.6 & 80.6 & 83.3 & 85.4 & 83.7 & 90.7 & 93.6 & 91.4 & 87.7 & 89.1 & 85.3$\pm$5.0 \\
    \cmidrule(lr){2-15}  
    & \textbf{\textit{Avg.}}    & \red{\underline{86.4}}$\pm$9.5 & 80.8$\pm$11. & 81.1$\pm$12. & 84.2$\pm$9.7 & 83.5$\pm$11. & 83.3$\pm$13. & 84.5$\pm$10. & \red{\textbf{88.6}}$\pm$8.9 & 86.1$\pm$12. & 84.6$\pm$14. & 85.5$\pm$12. & 86.0$\pm$11. & 84.6$\pm$11. \\
    \cmidrule(lr){2-15}  
    &  {\textit{p-value}}          & N/S &   N/S &   N/S &   N/S &   N/S &   N/S &   N/S & Ref. &  N/S &   N/S &   N/S &   N/S & - \\
        \cmidrule(lr){1-15}  
    & \textbf{\# Params.}  &  (513)  & (513)  & (514)  & (514)  & (514)  & (1026) & (0.07M)  & (0.33M)  & (0.4M)   &  (2.41M) &  (0.26M) &  (0.04M) &  ( - ) \\
        \cmidrule(lr){2-15}  
    \multirow{7}{*}{\rotatebox{90}{\makecell{ResNet18}}}
   & \textit{ImageNet}          & 88.4 & 70.4 & 83.0 & 82.7 & 82.6 & 86.1 & 88.1 & 87.7 & 87.1 & 89.9 & 88.5 & 88.7 & 85.3$\pm$5.1 \\
   \cmidrule{2-15}
    & Barlow T.                 & 95.6 & 90.3 & 93.4 & 93.5 & 91.1 & 96.8 & 95.5 & 95.8 & 95.3 & \underline{97.0} & 94.9 & 96.2 & 94.6$\pm$2.0 \\
    & MOCOv3                    & 95.3 & 91.4 & 93.9 & 92.3 & 92.2 & 95.1 & 94.7 & \underline{97.0} & 96.3 & 95.9 & \textbf{97.6} & 95.7 & 94.8$\pm$1.9 \\
    & SimCLR                    & 95.5 & 92.8 & 95.0 & 94.3 & 92.2 & 95.6 & 95.9 & 93.8 & 96.2 & 96.6 & 96.3 & 96.4 & 95.0$\pm$1.4 \\
    & BYOL                      & 94.8 & 91.1 & 91.6 & 91.2 & 91.8 & 95.9 & 93.7 & 94.6 & 96.0 & 95.9 & 95.0 & 95.7 & 93.9$\pm$1.9 \\
    & DINO                      & 94.3 & 89.9 & 92.8 & 92.4 & 91.3 & 95.7 & 95.2 & 95.2 & 95.4 & 95.3 & 96.5 & 95.9 & 94.2$\pm$2.0 \\
    \cmidrule(lr){2-15}  
    & \textbf{\textit{Avg.}}    & 95.1$\pm$0.5 & 91.1$\pm$1.0 & 93.3$\pm$1.1 & 92.7$\pm$1.1 & 91.7$\pm$0.5 & 95.8$\pm$0.6 & 95.0$\pm$0.8 & 95.3$\pm$1.1 & 95.8$\pm$0.4 & \red{\textbf{96.1}}$\pm$0.6 & \red{\underline{96.0}}$\pm$1.0 & 95.9$\pm$0.3 & 94.5$\pm$1.9 \\
    \cmidrule(lr){2-15}  
    &  {\textit{p-value}}          &   * &   * &   * &   * &   * &   N/S &   * &   N/S &   N/S & Ref. &  N/S &   N/S & - \\
    \cmidrule(lr){1-15}  
    & \textbf{\# Params.}  &  (2049) & (2049) & (2050) & (2050) & (2050) & (4098) & (0.26M)  & (4.46M)  & (1.18M)  &  (3.2M) &  (0.66M) &  (0.14M) &  ( - ) \\
        \cmidrule(lr){2-15}  
      \multirow{3}{*}{\rotatebox{90}{\makecell{ResNet50}}}
    & Barlow T.                 & 96.2 & 93.7 & 94.6 & 93.6 & 94.9 & 96.6 & 94.4 & 96.6 & 95.9 & 96.6 & 96.1 & 96.3 & 95.5$\pm$1.1 \\
    & MOCOv3                    & 96.8 & 93.2 & 94.3 & 95.3 & 94.1 & 96.4 & 96.5 & 97.2 & 97.2 & 95.7 & 95.9 & 96.6 & 95.8$\pm$1.2 \\
    & DINO                      & 95.6 & 94.3 & 95.9 & 93.1 & 93.5 & \underline{97.3} & 97.1 & 96.6 & 97.1 & \textbf{97.7} & 97.1 & 96.7 & 96.0$\pm$1.5 \\
    \cmidrule(lr){2-15}  
    & \textbf{\textit{Avg.}}    & 96.2$\pm$0.5 & 93.7$\pm$0.4 & 94.9$\pm$0.7 & 94.0$\pm$0.9 & 94.2$\pm$0.6 & \red{\textbf{96.8}}$\pm$0.4 & 96.0$\pm$1.2 & \red{\textbf{96.8}}$\pm$0.3 & \red{\underline{96.7}}$\pm$0.6 & \red{\underline{96.7}}$\pm$0.8 & 96.4$\pm$0.5 & 96.5$\pm$0.2 & 95.7$\pm$1.3 \\
    \cmidrule(lr){2-15}  
    &  {\textit{p-value}}          & N/S & * & * & * & * & Ref. & N/S & N/S & N/S & N/S & N/S & N/S & - \\
    \cmidrule(lr){1-15} 
    \end{tabular}}
    \label{tcga}
\end{table*}

\begin{table*}[]
    \centering
    \caption{{ Multi-class BRACS dataset at 10x. The first and second-best AUC scores are in bold and underlined respectively. The best average AUC score (per MIL) is in red. ImageNet-based results are not averaged.  {A Welch T-Test was applied to compare the best performing method (Ref.) with the others, where ($*$) indicates statistically significant differences ($p < 0.05$) and "N/S" denotes not significance. The \textit{DAMIL} method introduces additional complexity because the K-Means algorithm is applied independently on each slide.}}}
    \resizebox{\textwidth}{!}{
    \begin{tabular}{cccccccc|cccccc|c}
    \toprule
    
    \multirow{2}{*}{ \rotatebox{90}{\makecell{\textbf{Arch.}}}} &     
    \textbf{Pretrain.} &\textit{MeanMIL} & \textit{MaxMIL} & \textit{MixMIL} & \textit{AutoMIL} & \textit{LNPMIL} & \textit{AttenMIL} & \textit{ABMIL} & \textit{DSMIL} & \textit{CLAM} & \textit{TransMIL} &  {\textit{DTFDMIL}} &  {\textit{DAMIL}} &  {\textit{Avg.}} \\
    
    \cmidrule(lr){2-15}  
     & \textbf{Params.} & (1155) & (1155) & (1156) & (1156) & (1156) & (1540) & (0.05M)  & (0.33M)  & (0.2M)   & (2.34M) & (0.23M) & (0.03M) & (-) \\
    \cmidrule(lr){1-15}  
    \multirow{8}{*}{\rotatebox{90}{\makecell{ViT-Small}}} 
    & \textit{ImageNet}         & 68.9 & 60.0 & 66.8 & 73.4 & 71.7 & 80.0 & 75.9 & 72.8 & 75.6 & 80.9 & 75.5 & 63.0 & 72.0$\pm$6.1 \\
    \cmidrule{2-15}
    & Barlow T.                 & 72.9 & 69.0 & 72.1 & 75.1 & 75.2 & 78.6 & 74.8 & 77.6 & 79.9 & 80.7 & 84.4 & 85.2 & 77.1$\pm$4.7 \\
    & MOCOv3                    & 71.5 & 79.7 & 81.7 & 75.2 & 75.2 & 74.2 & 78.0 & 83.3 & 79.5 & \textbf{85.4} & 84.5 & 80.4 & 79.0$\pm$4.2 \\
    & SimCLR                    & 50.0 & 49.2 & 52.0 & 55.8 & 58.6 & 50.5 & 54.4 & 57.1 & 60.2 & 68.9 & 84.7 & 82.1 & 60.3$\pm$11.6 \\
    & BYOL                      & 71.3 & \textbf{85.4} & 81.7 & 76.7 & 81.0 & 75.5 & 84.9 & 84.2 & \underline{85.0} & 80.3 & 49.2 & 55.8 & 75.9$\pm$11.3 \\
    & DINO                      & 74.8 & 79.3 & 81.0 & 79.6 & 77.7 & 81.2 & 78.9 & 80.5 & 84.3 & 82.9 & 77.4 & 82.0 & 80.0$\pm$2.5 \\
    & MAE                       & 64.0 & 72.9 & 75.0 & 67.2 & 64.1 & 71.7 & 71.6 & 79.0 & 74.6 & 82.5 & 79.2 & 71.0 & 72.7$\pm$5.6 \\
    \cmidrule(lr){2-15}  
    & \textbf{\textit{Avg.}}    & 67.4$\pm$8.5 & 72.6$\pm$11.7 & 73.9$\pm$11. & 71.6$\pm$8.0 & 72.0$\pm$7.9 & 72.0$\pm$10. & 73.8$\pm$9.6 & 77.0$\pm$9.2 & \red{\underline{77.2}}$\pm$8.4 & \red{\textbf{80.1}}$\pm$5.3 & 76.6$\pm$12.6 & 76.1$\pm$10. & 74.2$\pm$10. \\
    \cmidrule(lr){2-15}  
    &  {\textit{p-value}}          & * &  N/S & N/S & N/S & N/S & N/S & N/S & N/S & N/S & Ref.& N/S & N/S & - \\
    \cmidrule(lr){1-15}  
    & \textbf{Params.}  & (1539) & (1539) & (1540) & (1540) & (1540) & (2052) & (0.07M)  & (0.33M)  & (0.4M)   & (2.41M) & (0.26M) & (0.04M) & (-) \\
    \cmidrule(lr){2-15}  
    \multirow{7}{*}{\rotatebox{90}{\makecell{ResNet18}}}
   & \textit{ImageNet}          & 71.6 & 70.3 & 60.0 & 68.8 & 69.2 & 77.5 & 78.0 & 77.3 & 76.4 & 76.1 & 80.5 & 72.8 & 73.2$\pm$5.4\\
   \cmidrule{2-15}
    & Barlow T.                 & 78.3 & 80.7 & 81.2 & 84.7 & 79.4 & 88.9 & 85.0 & 84.6 & 86.3 & 87.0 & 86.4 & 81.3 & 83.6$\pm$3.2 \\
    & MOCOv3                    & 76.0 & 82.9 & 85.0 & 82.2 & 77.2 & 86.1 & 84.4 & 85.3 & 85.7 & 86.3 & \underline{89.1} & 80.6 & 83.4$\pm$3.7 \\
    & SimCLR                    & 78.7 & 76.7 & 80.3 & 86.3 & 80.7 & 85.7 & 82.0 & 82.3 & 88.0 & 84.8 & 78.7 & 82.6 & 82.2$\pm$3.3 \\
    & BYOL                      & 77.1 & 79.5 & 84.5 & 79.7 & 76.1 & 86.4 & 85.4 & 85.3 & 86.3 & 83.8 & 84.7 & 84.6 & 82.8$\pm$3.5 \\
    & DINO                      & 75.7 & 79.0 & 79.8 & 79.2 & 80.4 & \textbf{89.4} & 88.5 & 83.6 & 88.8 & 86.5 & 88.2 & 79.9 & 83.2$\pm$4.6 \\
    \cmidrule(lr){2-15}  
    & \textbf{\textit{Avg.}}    & 77.2$\pm$1.2 & 79.8$\pm$2.0 & 82.2$\pm$2.2 & 82.4$\pm$2.8 & 78.8$\pm$1.8 & \red{\textbf{87.3}}$\pm$1.5 & 85.1$\pm$2.1 & 84.2$\pm$1.1 & \red{\underline{87.0}}$\pm$1.2 & 85.7$\pm$1.2 & 85.4$\pm$3.7 & 81.8$\pm$1.7 & 83.1$\pm$3.7 \\
    \cmidrule(lr){2-15}  
    &  {\textit{p-value}}          & * & * & * & * & * & Ref. & N/S & * & N/S & N/S & N/S & * & - \\
    \cmidrule(lr){1-15}
    \end{tabular}
    } 
    \label{bracs}
\end{table*}

\begin{table*}[]
    \centering
    \caption{{  VisioMel dataset at 10x. The first and second-best AUC scores are in bold and underlined respectively. The best average AUC score (per MIL) is in red. ImageNet-based results are not averaged.  {A Welch T-Test was applied to compare the best performing method (Ref.) with the others, where ($*$) indicates statistically significant differences ($p < 0.05$) and "N/S" denotes not significance. The \textit{DAMIL} method introduces additional complexity because the K-Means algorithm is applied independently on each slide.}}}
    \resizebox{\textwidth}{!}{
    \begin{tabular}{cccccccc|cccccc|c}
    \toprule
    \multirow{2}{*}{ \rotatebox{90}{\makecell{\textbf{Arch.}}}} &     
    \textbf{Pretrain.} &\textit{MeanMIL} & \textit{MaxMIL} & \textit{MixMIL} & \textit{AutoMIL} & \textit{LNPMIL} & \textit{AttenMIL} & \textit{ABMIL} & \textit{DSMIL} & \textit{CLAM} & \textit{TransMIL} &  {\textit{DTFDMIL}} &  {\textit{DAMIL}} &  {\textit{Avg.}} \\
    \cmidrule(lr){2-15}  
     & \textbf{Params.} &  (385)  & (385)  & (386)  & (386)  & (386)  & (770)  & (0.05M)  & (0.2M)   & (0.33M)  &  (2.34M) &  (0.23M) &  (0.03M) & (-) \\
    \cmidrule(lr){2-15}  
    \multirow{8}{*}{\rotatebox{90}{\makecell{ViT-Small}}} 
    & \textit{ImageNet}         & 69.2 & 73.9 & 71.5 & 74.3 & 52.0 & 48.5 & 59.9 & 75.3 & 60.9 & 72.9 & 70.7 & 67.9 & 66.4$\pm$8.6 \\
    \cmidrule{2-15}
    & Barlow T.                 & 74.1 & 70.7 & 73.2 & 72.4 & 59.6 & 68.0 & 71.4 & 72.9 & 55.3 & 69.9 & 71.5 & 62.3 & 68.4$\pm$5.8 \\
    & MOCOv3                    & 75.0 & 74.6 & 75.6 & 75.6 & 71.9 & 60.1 & 72.2 & 75.3 & 75.0 & 74.3 & 74.4 & 72.3 & 73.0$\pm$4.1 \\
    & SimCLR                    & 73.7 & 71.1 & 67.4 & 73.8 & 70.6 & 61.4 & 70.8 & 75.8 & 66.5 & 72.3 & 54.7 & 73.4 & 69.3$\pm$5.8 \\
    & BYOL                      & 71.5 & 68.1 & 73.1 & 72.2 & 68.2 & 51.3 & 70.5 & \textbf{77.5} & 65.7 & 64.4 & 57.6 & 68.6 & 67.4$\pm$6.8 \\
    & DINO                      & \underline{76.2} & 69.6 & 70.0 & 74.6 & 66.8 & 72.1 & 68.7 & 75.3 & 63.4 & 70.9 & 73.8 & 73.3 & 71.2$\pm$3.6 \\
    & MAE                       & 70.1 & 69.6 & 71.9 & 71.4 & 56.5 & 47.0 & 61.6 & 70.7 & 67.2 & 60.7 & 68.1 & 65.6 & 65.0$\pm$7.1 \\
    \cmidrule(lr){2-15}  
    & \textbf{\textit{Avg.}}    & \red{\underline{73.4}}$\pm$2.1 & 70.6$\pm$2.0 & 71.9$\pm$2.6 & 73.3$\pm$1.5 & 65.6$\pm$5.7 & 60.0$\pm$8.7 & 69.2$\pm$3.6 & \red{\textbf{74.6}}$\pm$2.2 & 65.5$\pm$5.8 & 68.7$\pm$4.7 & 66.7$\pm$7.8 & 69.2$\pm$4.2 & 69.1$\pm$6.2 \\
    \cmidrule(lr){2-15}  
    &  {\textit{p-value}}           & N/S & * & N/S & N/S & * & * & * & Ref. & * & * & N/S & * & - \\
        \cmidrule(lr){1-15}  
    & \textbf{Params.}  &  (513)  & (513)  & (514)  & (514)  & (514)  & (1026) & (0.07M)  & (0.33M)  & (0.4M)   &  (2.41M)  &  (0.18M)  &  (0.04M) & (-) \\
        \cmidrule(lr){2-15}  
    
    \multirow{7}{*}{\rotatebox{90}{\makecell{ResNet18}}}
    & \textit{ImageNet}         & 74.2 & 66.1 & 73.0 & 69.4 & 50.9 & 52.9 & 70.6 & 76.1 & 67.5 & 61.4 & 67.2 & 70.1 & 66.6$\pm$7.6 \\
    \cmidrule{2-15}
    & Barlow T.                 & 75.3 & 67.9 & 72.7 & 72.5 & 62.7 & 75.2 & 67.7 & 69.7 & 68.0 & 75.7 & 61.6 & 68.2 & 69.8$\pm$4.5 \\
    & MOCOv3                    & \underline{76.4} & 75.3 & 73.2 & 73.8 & 53.7 & 72.3 & 75.3 & 71.4 & 67.5 & 68.0 & 61.6 & 65.4 & 69.5$\pm$6.4 \\
    & SimCLR                    & 73.9 & 69.6 & 74.3 & 72.9 & 69.2 & 75.3 & 63.9 & 68.6 & 70.0 & 65.3 & 65.7 & 73.6 & 70.2$\pm$3.7 \\
    & BYOL                      & 73.9 & 72.0 & 71.7 & 64.5 & 73.2 & 44.3 & 71.1 & 67.4 & 68.4 & 64.6 & \textbf{78.8} & 74.2 & 68.7$\pm$8.4 \\
    & DINO                      & 75.5 & 67.1 & 73.8 & 69.9 & 73.2 & 73.9 & 64.5 & 72.7 & 73.4 & 65.2 & 61.1 & 73.0 & 70.3$\pm$4.5 \\
    \cmidrule(lr){2-15}  
    & \textbf{\textit{Avg.}}    & \red{\textbf{75.0}}$\pm$1.0 & 70.4$\pm$3.0 & \red{\underline{73.1}}$\pm$0.9 & 70.7$\pm$3.4 & 66.4$\pm$7.4 & 68.2$\pm$12. & 68.5$\pm$4.3 & 70.0$\pm$1.9 & 69.5$\pm$2.1 & 67.8$\pm$4.1 & 65.8$\pm$6.7 & 70.9$\pm$3.5 & 69.7$\pm$5.8 \\
    \cmidrule(lr){2-15}  
    &  {\textit{p-value}}          & Ref. &  * &   * &   * &   N/S &   N/S &   * &   * &   * &   * &   * &   N/S & - \\
    \cmidrule(lr){1-15}  
    \end{tabular} } 
    \label{mel}
\end{table*}

\subsection{Discussion}
In this section, we discuss the results described in the previous section. We highlight various important questions that we believe are particularly relevant to the histopathology community and provide detailed answers based on the obtained results and analyses. \\
\noindent \textit{\textbf{Do we need complex embedding-based MIL? }} No. Based on our results  {(see Fig. \ref{fig:rocs})}, we can notice that instance-based MIL methods, coupled with good SSL feature extractors, are either the best or the second best performing methods, both in terms of combinations and averaged AUC scores, across all four datasets.\\
\textit{\textbf{ What is the best backbone? }} CNNs seem more robust than VITs. Furthermore, the performance improves with larger backbones: the best AUC with ResNet50 on Camelyon16 is 3 AUC points above the best AUC with ResNet18. The same trend is observed for TCGA-NSCLC. However, the answer also depends on the computational capability and the magnification level of the data (and thus number of samples). With great computational capability, one should use data at 20x and big architectures, such as ResNet50 or ViT\_Large. However, with limited computational power or data at only 10x, one should use ResNet18.\\ 
\textit{\textbf{ What is the best SSL? }} Our best results are obtained with SSL pre-training, rather than ImageNet initialization. Among SSL methods, results may vary and the best results are obtained using either MOCOv3, DINO or Barlow Twins, with the latter being particularly advantageous due to its simple architecture and ease of tuning. We can notice that SimCLR, as already shown in the literature, falls short in terms of performance.\\
\textit{\textbf{Do we benefit from pathology-specific SSL techniques?}} Pathology-adapted techniques improve the performance of traditional SSL methods. Using specific augmentations and SSL methods for pathology helps achieve better results. Since SSL methods are very sensitive to augmentations, using domain knowledge to adapt these augmentations greatly enhances generalization. For example, CluBYOL, the pathology-adapted version of the BYOL framework, significantly outperforms it by 8.6 AUC points, averaged across all MIL methods, on the Camelyon16 dataset at 10x magnification. Furthermore, CluBYOL{\scriptsize path} achieves an even greater enhancement, adding an additional 1.8 AUC points, resulting in a total improvement of 10.4 AUC points (see Table \ref{path}).\\
\textit{\textbf{Can we leverage foundation models?}}
Yes, we can use foundation models as feature extractors, but it's important to use those specifically adapted to pathology. As highlighted in the previous section, there is a significant performance gap between DINOv2 and the other models, underscoring the critical role of in-domain knowledge. Unlike the other models, DINOv2 was not trained on pathological images, which emphasizes how pre-training on relevant, domain-specific data can greatly enhance performance. Among the foundation models, UNI achieves the highest performance, likely due to several factors. It benefits from having the largest and most diverse dataset, consisting of over 100 million patches, which is 5-6 times larger than others. Additionally, UNI uses the latest SSL framework and larger backbone architecture.\\
\textit{\textbf{Should we simply use MaxMIL?}} Even if MaxMIL can give the best AUC score (see Table \ref{bracs}), it should not always be chosen.  {As shown in Fig. \ref{fig:rocs}, the method does not always achieve the best results.} In particular, we can notice that in Table \ref{tcga} and \ref{mel}, the best results are obtained using MeanMIL. This is because in TCGA-NSCLC and VisioMel tumor sizes are larger, typically covering more than 80\% of tissue regions. Thus, averaging scores across several patches improves the result. Nevertheless, when tumor size is unknown or variable and pathology markers can be subtle, one should use an in-between approach, such as the newly proposed instance-based MIL methods.
\subsection{Limitations and Perspectives}
\noindent The analysis presented in this paper has some limitations that we discuss hereafter:\\
\textbf{ \textit{Multiple magnification levels.}} We did not use  multiple magnification levels due to computational complexity. Extracting patches at different magnification levels significantly increases the dataset size, approximately 4-5 times for each higher resolution level, making it very expensive for our computational resources. However, as noted by \cite{chen_scaling_2022} and \cite{kang_benchmarking_2023}, this approach improves the overall model performance, probably because it mimics pathologists, who typically examine tissues at various magnifications when making diagnostic decisions. \\       
\textit{\textbf{Other type of MIL.}} Here, we included in our analysis 4 embedding-based MIL methods which recently achieved SOTA results. However, other MIL methods exist which leverage, for instance, multi-levels of magnification \cite{deng2024cross, chen_scaling_2022, qu_boosting_2023} or graph structures \cite{li_graph_2018, guan_node-aligned_2022}, and it would be of interest to include them in our comparison.\\
\textit{\textbf{Influence of number of training samples $\mathbf{N}$.}} Instance-based MIL methods, in particular MaxMIL, theoretically require more WSIs to work, since only a small number of instances within each slide (1 in MaxMIL) can actually participate in the decision \cite{shao_transmil_2021}. At the same time, embedding-based MIL methods have more parameters and thus require more training data. In the future, it would be interesting to understand the impact of $N$ on both classes of MIL methods.\\
\textit{\textbf{Better pathology-adapted SSL methods.}} The current pathology-adapted SSL methods often involve minor modifications compared to standard SSL methods originally designed for natural images, such as adjustments in augmentation techniques, positive/negative sampling, or clustering strategies. More effort should be put into developing well-adapted SSL methods that leverage not only the different magnification levels but also the prior medical knowledge of the pathologists. An SSL approach that mimics the reasoning process of a pathologist could potentially be integrated into foundation models like UNI, enhancing their applicability and performance in WSI analysis tasks.

\section{Conclusions}
 In this paper, we conducted a large-scale study using 6 SSL with 4 backbones, 4 foundation models and 10 MIL methods on four diverse datasets, covering binary and multi-class classification tasks with increasing clinical complexity. Our results demonstrate that simple instance-based MIL methods with very few parameters, combined with robust SSL feature extractors, can achieve competitive or superior performance than complex embedding-based MIL methods across different backbones. The newly proposed instance-based MIL methods achieve new SOTA results on BRACS and Camelyon16. By sharing our code, pre-trained models, and insights, we aim to provide valuable resources for future research in this domain.

\section{Code and Data Availability}
The code for implementing this pipeline is available on GitHub at \url{https://github.com/mammadov7/wsi_classification_pipeline}.
In this study, we used four publicly available datasets:
\begin{itemize}
    \item \textbf{VisioMel} - melanoma slides data, accessible at \url{https://www.drivendata.org/competitions/148/visiomel-melanoma/data/} 
    \item \textbf{TCGA-NSCLC} - non-small cell lung cancer data, obtained from the TCGA-LUAD and TCGA-LUSC projects, available via \url{https://portal.gdc.cancer.gov/analysis_page?app=Projects}
    \item \textbf{BRACS} - breast cancer images available at \url{https://www.bracs.icar.cnr.it/}
    \item \textbf{Camelyon16} -  lymph node metastasis data, accessible at \url{http://gigadb.org/dataset/100439}.
\end{itemize}

\section{Acknowledgements}
This paper has been supported by the French National Research Agency (ANR-20-THIA-0012) and by the Hi!PARIS Center on Data Analytics and Artificial Intelligence. Furthermore, this work was performed using HPC resources from GENCI-IDRIS (Grant 2023-AD011013982R1).

\bibliography{report}

\end{document}